\documentclass[onefignum,onetabnum]{siamart220329}

\usepackage{lipsum}
\usepackage{amsfonts,amssymb}
\usepackage{graphicx}
\usepackage{epstopdf}
\usepackage{algorithmic}
\usepackage{comment}

\ifpdf
  \DeclareGraphicsExtensions{.eps,.pdf,.png,.jpg}
\else
  \DeclareGraphicsExtensions{.eps}
\fi

\definecolor{red2}{RGB}{204,0,0}
\definecolor{blue2}{RGB}{0,103,165}
\hypersetup{colorlinks,linkcolor=[RGB]{0,103,165},citecolor=[RGB]{180,0,0}}

\newsiamremark{remark}{Remark}
\newsiamremark{hypothesis}{Hypothesis}
\crefname{hypothesis}{Hypothesis}{Hypotheses}
\newsiamthm{claim}{Claim}

\def\be{\begin{equation}}
\def\ee{\end{equation}}

\def\x{\mathbf{x}}
\def\xt{\widetilde{\x}}
\def\y{\mathbf{y}}

\def\z{\mathbf{z}}

\def\xt{\widetilde{\x}}

\def\Rs{\mathbb{R}}

\def\G{\mathbf{G}}
\def\Gt{\mathbf{\widetilde{G}}}

\def\Pi{\mathbf{\Phi}}

\def\D{\mathbf{D}}
\def\E{\mathbf{E}}
\def\S{\mathbf{S}}

\newcommand{\norm}[1]{\left\lVert#1\right\rVert}

\headers{Learning SDE via Autoencoders}{Z. Xu, Y. Chen, Q. Chen and D.Xiu}

\title{Modeling Unknown Stochastic Dynamical System via Autoencoder}

\author{Zhongshu Xu, Yuan Chen, Qifan Chen and Dongbin Xiu\thanks{E-mail addresses: \texttt{\{xu.4202,chen.11050,chen.11010,xiu.16\}@osu.edu}. Department of Mathematics, The Ohio State University, Columbus, OH 43210, USA. Funding: This work was partially supported by AFOSR FA9550-22-1-0011.}}

\usepackage{amsopn}

\ifpdf
\hypersetup{
  pdftitle={Title},
  pdfauthor={Yuan Chen}
}
\fi

\begin{document}

\maketitle

\begin{abstract}
We present a numerical method to learn an accurate predictive model for an unknown stochastic dynamical system from its trajectory data. The method seeks to approximate the unknown flow map of the underlying system. It employs the idea of autoencoder to identify the unobserved latent random variables. In our approach, we design an encoding function to discover the latent variables, which are modeled as unit Gaussian, and a decoding function to reconstruct the system's future states. Both the encoder and decoder are expressed as deep neural networks (DNNs). Once the DNNs are trained by the trajectory data, the decoder serves as a predictive model for the unknown stochastic system. Through an extensive set of numerical examples, we demonstrate that the method is able to produce long-term system predictions by using short bursts of trajectory data. It is also applicable to systems driven by non-Gaussian noises.
\end{abstract}



\section{Introduction}
Designing data-driven methods to discover unknown physical systems has attracted an increasing amount of attention recently. The goal is to discover the fundamental laws or equations behind the measurement data, in order to construct an effective predictive model for the unknown dynamics. Most of the existing methods are developed for learning deterministic dynamical systems. These include SINDy (\cite{brunton2016discovering}), physics-informed neural networks (PINNs) (\cite{raissi2019physics,raissi2018multistep}), Fourier neural operator (FNO) (\cite{li2020fourier}), computational graph completion (\cite{owhadi2021computational}), sparsity promoting methods (\cite{schaeffer2017sparse,schaeffer2018extracting,kang2019identifying}), flow map learning (FML) (\cite{qin2019data, Churchill_2023}), to name a few.

For a stochastic system, noises in data, along with the unobservability of the inherent stochasticity in the system, pose significant challenges in learning the system.
Most of the existing methods focus on learning It\^o type stochastic differential equations (SDEs). These methods employ techniques such as Gaussian process (\cite{yildiz2018learning,pmlr-v1-archambeau07a,darcy2022one,opper2019variational}), polynomial approximations (\cite{wang2022data, li2021data}), deep neural networks (DNNs) \cite{chen2023data,yang2022generative,chen2021solving,zhang2022multiauto}, etc. More recently, a stochastic extension of the deterministic FML approach (\cite{qin2019data, Churchill_2023}) was proposed in \cite{chen2023learning}, where generative model such as GANs (generative adversarial networks) was employed.

The focus, as well as the contribution, of this paper, is on the development of a new generative model for data-driven modeling of unknown stochastic dynamical systems. Similar to the work of \cite{chen2023learning}, the new method also employs the broad framework of FML, where a time-stepper is constructed using short bursts of measurement data to provide reliable long-term predictions of the unknown system. The use of GANs in \cite{chen2023learning}, albeit effective, is computationally challenging. This is because, by its mathematical formulation, GANs are intrinsically difficult to train to high accuracy, which is critical for long-term system prediction. To circumvent to the computational difficulty, we adopt the concept of autoencoder and propose a novel stochastic FML (sFML) for unknown SDEs. Autoencoder is a type of DNN used to learn effective codings of unlabelled data and is widely used in problems such as classification \cite{goodfellow2016deep}, principal component analysis (PCA) \cite{hinton2006reducing}, image processing \cite{theis2017lossy,buades2005review}, etc. It has also shown promises in scientific computing for problems including operator learning \cite{oommen2022learning}, forward and backward SDEs \cite{zhong2023pi, zhang2022multiauto, hasan2021identifying}, uncertainty quantification \cite{liu2022uncertainty}, etc. A standard autoencoder learns two functions: an encoding function that learns the latent variables of the input data, and a decoding function that recreates the input data from the encoded representation. In the proposed method, we design an encoding function to extract the hidden latent random variables from the noisy trajectory data of the unknown SDE, and a decoding function to reconstruct the future state data given the current state data and the latent random variables learned from the encoder. Both the encoder and the decoder are expressed as DNNs. The training of the encoder is carried out by enforcing the latent variables to be unit Gaussian independent of the current state data. By doing so, we effectively assume that the unknown SDEs can be expressed, or approximated, by an It\^o type SDE driven by a Wiener process. The training loss of the entire sFML autoencoder consists of two parts: a reconstruction loss for the decoder and a distributional loss for the encoder to enforce the latent variables to be unit Gaussian. An important sub-sampling strategy of the training data is also developed to ensure that the latent variables are independent of the current state variables.  Upon successful training of the entire DNN autoencoder, the decoder becomes a predictive model for long-term system predictions. An extensive set of numerical examples, including learning unknown SDEs with non-Gaussian noises, are presented to demonstrate the effectiveness of the proposed autoencoder sFML approach. Our code is accessible at \url{https://github.com/AtticusXu/Modeling-Unknown-Stochastic-Dynamical-System-via-Autoencoder}

\section{Setup and preliminaries}
\label{sec:setup}

Let us consider a $d$-dimensional ($d\geq 1$) stochastic process, $ \x_t := \x(\omega, t): \Omega \times [0,T] \mapsto \mathbb{R}^d$, $d \geq 1$, driven by an unknown stochastic differential equation (SDE), where $\Omega$ is the event space and $T>0$ a finite time horizon. We are interested in constructing an accurate numerical model for the unknown SDE by using measurement data of $\x_t$.

\subsection{Assumption}
We assume the stochastic process $\x_t$ is time-homogenous, (cf. \cite{oksendal2003stochastic}, Chapter 7), in the sense that for any time $\Delta\geq 0$, 
\begin{equation} \label{autonomous}
\mathbb{P}(\mathbf{x}_{s+\Delta}|\mathbf{x}_{s}=\x)=\mathbb{P}(\mathbf{x}_{\Delta}|\mathbf{x}_{0}=\x),\qquad s\geq 0.
\end{equation}
More specifically, $\x_t$  is a time-homogeneous It\^o diffusion driven by
\be \label{Ito0}
d\x_t = \mathbf{b}(\x_t)dt + \sigma(\x_t) d\mathbf{W}_t,
\ee
where $\mathbf{W}_t$ is $m$-dimensional Brownian motion, and $\mathbf{b}:\Rs^d\to\Rs^d$, $\sigma:\Rs^d\to\Rs^{d\times m}$, $m\geq 1$, satisfy appropriate conditions (e.g., Lipschitz continuity).
%
However, we assume that no information about the equation \eqref{Ito0} is available: $\mathbf{b}$ and $\sigma$ are not known, the  Brownian motion $\mathbf{W}_t$ is not observable, and even its dimension $m$  is not known.

\subsection{Data}
We assume that $N_T \geq 1$ solution trajectories of $\x_t$ are observed over discrete time instances,
\be \label{traj0}
    \x\left(t_0^{(i)}\right),\x\left(t_1^{(i)}\right),\dots,\x\left(t_{L_i}^{(i)}\right), \qquad i=1,...,N_T,
\ee
where $(L_i+1)$ is the length of the $i$-th observation sequence. For simplicity, we assume a constant time lag among the time instances and trajectory length, i.e. $t^{(i)}_n-t^{(i)}_{n-1}\equiv \Delta $, for any $n, i \geq 1$, and $L_i \equiv L \geq 1$, for all $i$.

In the proposed modeling approach, the time instances are not required, thanks to the time homogeneity condition. We therefore assume the trajectory data take the following form
\be \label{traj}
    \x_0^{(i)},\x_1^{(i)},\dots,\x_{L}^{(i)}, \qquad i=1,...,N_T,
\ee
which follows the same format as in \eqref{traj0}, except the time variables are not required. 

\subsection{Objective and Related Work}

Our goal is to use the trajectory data \eqref{traj} to
construct an iterative predictive model
\be \label{model}
  \xt_{n+1} = \G_\Delta (\xt_n; \z_n), \qquad \z_n\in\Rs^{n_z}, \quad
  n\geq 0,
\ee
where $\z_n$ is a $n_z$-dimensional standard random vector with $n_z\geq 1$. When given an initial condition $\xt_0 = \x_0$,  the model \eqref{model} produces a trajectory that approximates the true trajectory in distribution, i.e.,
\be \label{weak}
  (\xt_0, \xt_1, \dots, \xt_N) \stackrel{d}{\approx} (\x_0, \x_1, \dots, \x_N),
\ee
for any finite $N$. 

The numerical model \eqref{model} takes the form of a time stepper. It follows the data-driven modeling framework of flow map learning (FML) that was first proposed in \cite{qin2019data} for deterministic dynamical systems. (See \cite{Churchill_2023} for a review.) For modeling unknown stochastic systems, the work of \cite{chen2023learning} proposed a stochastic flow map learning (sFML) model
$$
  \G_\Delta(\x; \z) = \D_\Delta(\x) + \S_\Delta(\x; \z),
$$
where $\D_\Delta$ is a deterministic sub-map and $\S_\Delta$ a stochastic sub-map. The deterministic sub-map approximates the conditional mean of the system, and the stochastic sub-map takes the form of generated adversarial networks (GANs) to accomplish the approximation goal \eqref{weak}. (See \cite{chen2023learning} for details.) Though effective, GANs are known to be difficult to train to high accuracy, which is critical for achieving long-term predictive accuracy of the sFML model \eqref{model}.

\subsection{Contribution}

In this paper, we propose a new construction of the sFML predictive model \eqref{model} that offers more robust DNN training and better accuracy control. This is accomplished by employing the concept of autoencoder. A standard autoencoder learns two functions: an encoding function that transforms the input data, and a decoding function that recreates the input data from the encoded representation. In our work, we focus on the training data set \eqref{traj} and choose adjacent pairs $(\x_n, \x_{n+1})$. We then train an encoding function to identify the unobserved (hidden) stochastic component:
$$
(\x_n, \x_{n+1}) \rightarrow \z_n,
$$
and a decoding function to reconstruct the trajectory
$$
(\x_n, \z_n) \rightarrow \xt_{n+1},
$$
such that $\xt_{n+1} \approx \x_{n+1}$.

By properly imposing conditions on the latent stochastic component $\z_n$ and defining a training loss function, we demonstrate that the autoencoder sFML approach can yield a highly effective predictive model \eqref{model}. Compared with the approach in \cite{chen2023learning}, the current autodecoder sFML approach can also effectively discover the true dimensions of unobserved stochastic component of the system via the learning of the latent variable $\z_n$. This provides a more detailed and quantitative understanding of the unknown stochastic dynamical system. Moreover, our extensive numerical experimentations suggest that the autoencoder sFML approach allows more robust DNN training and yields higher accuracy in the corresponding predictive models.


\section{Autoencoder Stochastic Flow Map Learning}
\label{sec:method}

In this section, we describe the details of the proposed autoencoder sFML approach for modeling unknown stochastic dynamical systems. We first present the mathematical motivation, then discuss the numerical algorithm, particularly the DNN structure and loss function, followed by a discussion of several important numerical implementation details.

\subsection{Mathematical Motivation}
\label{sec:math}

For the training data set \eqref{traj}, we consider its data pairs, separated by the constant time step $\Delta$,
$$
(\x_0^{(i)}, \x_1^{(i)}), \quad (\x_1^{(i)}, \x_2^{(i)}), \quad \dots, \quad
(\x_{L-1}^{(i)}, \x_L^{(i)}), \qquad i=1,\dots, N_T,
$$
which contain a total of $M=N_T L$ such data pairs. Since the stochastic process is time-homogeneous, we rename each pair using indices 0 and 1, and re-organize the training data set into a set of such pairwise data entries, 
\be \label{data_set}
\left(\x_0^{(i)}, \x_{1}^{(i)}\right), \qquad i=1,\dots, M, \qquad M=N_T L.
\ee
In other words, the training data set is re-arranged into $M$ numbers of very short trajectories of only two entries. Each of the $i$-th trajectory, $i=1,\dots, M$, starts with an ``initial condition'' $\x_0^{(i)}$ and ends a single time step $\Delta$ later at $\x_1^{(i)}$.

Since the process $\x_t$ follows the (unknown) time-homogeneous Ito diffusion \eqref{Ito0}, we have
\be 
\x_1 = \x_0 + \int_0^\Delta \mathbf{b}(\x_s) ds + \int_0^\Delta
\sigma(\x_s) d\mathbf{W}_s, 
\ee
where the training data \eqref{data_set} can be considered as sample paths of the process. This is the basis of our modeling principle: Given the training data \eqref{data_set}, there exists a standard Gaussian random variable $\z\sim \mathcal{N}(0, \mathbf{I}_{n_z})$, $n_z\geq 1$, independent of $\x_0$, and a function $\G_\Delta$ such that
\be
\x_1^{(i)} = \x_0^{(i)} + \G_\Delta(\x_0^{(i)}, \z^{(i)}), \qquad
i=1,\dots, M.
\ee

Here, the standard Gaussian random variable $\z$ is the ``hidden'' latent variable we seek to learn via the encoding function. It represents the increment jump of the Brownian motion $\mathbf{W}_t$ over $\Delta$, which follows a Gaussian distribution. Since the dimension of $\mathbf{W}_t$ is assumed to be unknown, the ``true'' dimension of $\z$, $n_z$, is also unknown.

Once the hidden variable $\z$ is learned by the encoding function, the unknown function $\G_\Delta$ will be learned by a decoding function. This is the design principle of our autodecoder sFML modeling of the SDE behind the data $\x_t$.

\subsection{Method Description}

Our autoencoder sFML method consists of two functions: an encoding function $\E_\Delta$ and a decoding function $\D_\Delta$:
\be \label{AEfunc}
\begin{split}
  &\E_\Delta: \mathbb{R}^d \times \mathbb{R}^d \mapsto
  \mathbb{R}^{n_{\z}}, \quad \textrm{(Encoder)}\\
  &\D_\Delta: \mathbb{R}^d \times \mathbb{R}^{n_{\z}} \mapsto
  \mathbb{R}^d. \quad \textrm{(Decoder)}
  \end{split}
\ee
Both functions are realized via DNNs, which are to be trained via the training data set \eqref{data_set} with properly defined loss functions.

\subsubsection{Encoding Function}
\label{sec:encode}

The encoding function $\E_\Delta$ in \eqref{AEfunc} takes a data
pair from the training data set \eqref{data_set} and returns a random vector of dimension $n_z$, 
\be \label{encode}
\z^{(i)} = \E_\Delta(\x_0^{(i)}, \x_1^{(i)}), \qquad i=1,\dots, M.
\ee
The goal is to enforce two conditions: (i) $\z \sim \mathcal{N}(0, \mathbf{I}_{n_z})$, unit Gaussian distribution of dimension $n_z$, where $\mathbf{I}_{n_z}$ is the identity matrix of size $n_z\times n_z$; and (ii) $\z$ is independent of $\x_0$.
Since both conditions are on the probabilistic properties of $\z$, we consider sets of the output of the encoder. Specifically, we divide the entire output set $\{\z^{(i)}, i=1,\dots,M\}$ of $M$ samples into $n_B>1$ ``batches'' of $N$ samples with $n_B N = M$. (For notational simplicity, we assume each batch contains an equal $N$ number of samples.)

Let us consider one batch of $N$ samples, $B = \left\{\z^{(i)}, i=1,\dots, N\right\}.$
We seek to enforce the distribution of the batch to be $n_z$-dimensional standard unit Gaussian distribution by minimizing a distributional loss function:
\be \label{LossD}
    \mathcal{L}_D(B) =  \mathcal{L}_{\textrm{distance}}(B) + \tau \cdot \mathcal{L}_{\textrm{moment}}(B),
\ee
where $\mathcal{L}_{\textrm{distance}}$ measures a statistical distance between the distribution of the batch $B$ and the distribution of $n_z$-dimensional standard Gaussian, and $\mathcal{L}_{\textrm{moment}}$ measures the deviation of the moments of $B$ to those of $n_z$-dimensional standard Gaussian, and $\tau>0$ is a penalty parameter. Through our extensive numerical experimentation, we found that Renyi-entropy-based distance is suitable for $\mathcal{L}_{\textrm{distance}}$, and it is necessary to include the moment loss $\mathcal{L}_{\textrm{moment}}$ to ensure Gaussianity of the latent variable $\z$. The technical details of $\mathcal{L}_{\textrm{distance}}$ and $\mathcal{L}_{\textrm{moment}}$ are in Section \ref{sec:statdis} and \ref{sec:moment}, respectively.

While the minimization of the distributional loss \eqref{LossD} can force the batch samples to follow the unit normal distribution, it is not sufficient to ensure the samples are independent of $\x_0$. To enforce the independence condition, we recognize that the samples $\z^{(i)}$ in the batch $B$ are computed by the encoder via \eqref{encode}. Consequently, sampling $\z^{(i)}$ is determined by sampling $\x_0^{(i)}$ in the training data set
\eqref{data_set} via
\be \label{batch}
 B = \left\{\z^{(i)} = \E_\Delta\left(\x_0^{(i)}, \x_1^{(i)}\right)\right\}_{i=1}^N,
 \ee
Therefore, to enforce the samples $\z^{(i)}$ in the batch $B$ to have the standard unit Gaussian distribution {\em and} be independent of $\x_0$, we propose the following sub-sampling principle for the encoding function construction:
\begin{quote}
  For each batch $B$ \eqref{batch}, let $\left\{ \x_0^{(i)}\right\}_{i=1}^N$ be sampled from an
  arbitrary non-Gaussian distribution out of the training data set
  \eqref{data_set}, the encoding function \eqref{encode} is determined
  by minimizing the distributional loss function \eqref{LossD} over
  the batch $B$.
\end{quote}
The principle is applied to each of the $n_B>1$ batches. This ensures that $\z$ follows the standard unit Gaussian distribution regardless of the distribution of $\x_0$. Subsequently, this promotes independence between $\z$ and $\x_0$. The effective way to ``arbitrarily'' sample $\x_0$ is discussed in detail in Section \ref{sec:resample}.
\begin{remark}
The proposed principle is critical in promoting independence between $\z$ and $\x_0$. Consider, for example, a case when the unknown SDE follows a simple It\^o SDE \eqref{Ito0} with a constant diffusion. In this case, if $\{\x_0^{(i)}\}$ are sampled following a Gaussian distribution, then $\{\x_1^{(i)}\}$ will follow a Gaussian distribution as well. Consequently, the encoding function $\E_\Delta$ can be defined as a linear combination of $\x_0$ and $\x_1$ to return a standard Gaussian variable $\z$. Such an encoder would satisfy a successful minimization of the distributional loss \eqref{LossD} but produce a latent variable $\z$ that is dependent on $\x_0$.
\end{remark}

\subsubsection{Decoding Function}

The decoding function $\D_\Delta$ in \eqref{AEfunc} seeks to replicate $\x_1$ by taking inputs from $\x_0$ and the latent standard normal variable $\z$ identified by the encoder. Specifically, consider the same batch samples $B$ in \eqref{batch}. We define the decoding function as
\be \label{decode}
 \xt^{(i)}_1 = \D_\Delta(\x_0^{(i)}, \z^{(i)}), \qquad i=1,\dots, N.
\ee
The decoder is then constructed by minimizing mean squared error (MSE) loss function
\be \label{Loss2}
    \mathcal{L}_{MSE}(B)= \frac{1}{N}\sum_{i=1}^{N} \left\|\x_1^{(i)} - \xt^{(i)}_1 \right\|_2^2.
\ee

\subsubsection{DNN Structure and Loss Function}

\begin{figure}[htbp]
  \centering
  \includegraphics[width=.8\textwidth]{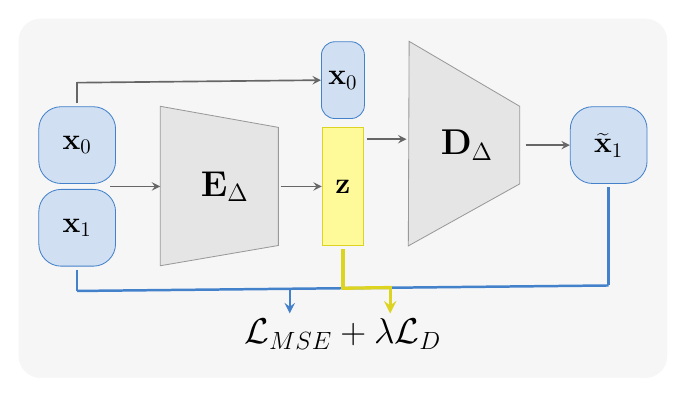}
  \caption{An illustration of the network structure and training loss for the proposed autoencoder sFML method.}
  \label{fig:AEnet}
\end{figure}

Both the encoder \eqref{encode} and decoder \eqref{decode} are
constructed using DNNs. In this paper, they both take the form of standard fully connected feedforward networks. The complete DNN structure is shown in Figure \ref{fig:AEnet}. The encoder takes the inputs of $\x_0$ and $\x_1$ from the training data set
\eqref{data_set} and produces the output $\z$; the decoder takes $\z$, along with $\x_0$ and produces the output $\xt_1$. The training of the DNN is conducted by minimizing the following loss function averaged over the
$n_B$ batches:
\be
\mathcal{L} = \frac{1}{n_B} \sum_{j=1}^{n_B} \mathcal{L}(B_j)=
\frac{1}{n_B} \sum_{j=1}^{n_B} \left[ 
\mathcal{L}_{MSE} (B_j) + \lambda \cdot \mathcal{L}_D(B_j)\right],
\ee
where $\mathcal{L}_D$ is the distributional loss \eqref{LossD}, $\mathcal{L}_{MSE}$ the MSE loss \eqref{Loss2}, and $\lambda>0$ a scaling parameter.

\subsection{Algorithm Detail}

In this section, we discuss a few important technical details for the implementation of the proposed autoencoder sFML method.

\subsubsection{Sub-sampling Strategy}
\label{sec:resample}

\begin{figure}[htbp]
  \centering
  \includegraphics[width=.98\textwidth]{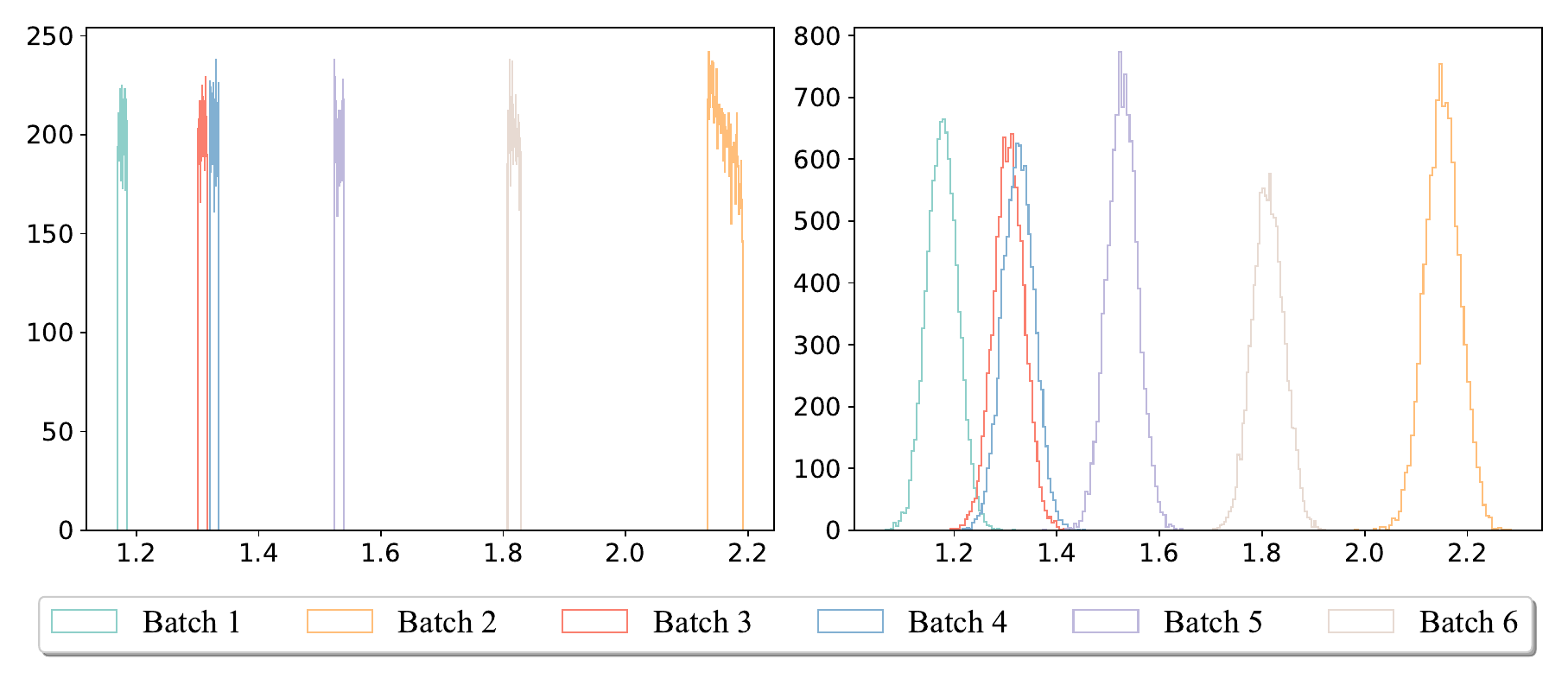}
  \caption{An illustration of the proposed batch sampling: Left:
    histograms of 6 batches of sampled $\x_0$'s; Right: histograms of the
    corresponding $\x_1$'s.}
  \label{fig:resample}
\end{figure}

As discussed in Section \ref{sec:encode}, it is critical to select each of the $n_B$ batches by sampling $\x_0$ in an arbitrary way out of the training data set \eqref{data_set}. To accomplish this, we propose the following sampling strategy:
consider all the samples of the entire training data set \eqref{data_set}
$$
\left\{\x_0^{(i)}, i=1,\dots, M\right\}.
$$
(Since $\x_0$ and $\x_1$ appear in pairs, samples of $\x_1$ are immediately determined once samples of $\x_0$ are chosen.) To construct $n_B>1$ batches, each of which contains $N$ samples ($M=N n_B$), we proceed as follows:
\begin{itemize}
    \item Randomly choose $n_B$ samples of $\x_0$ from \eqref{data_set} using uniform distribution;
    \item For each of the chosen $\x_0$ samples, find its $(N-1)$
      nearest neighbor points to form a batch with $N$ samples.
\end{itemize}
The procedure is carried out at the beginning of each epoch. This is to ensure the ``arbitrariness'' of the distribution of $\x_0$  in the batches. An illustrative example of sampling of 6 batches is shown in Figure \ref{fig:resample}. The distribution of the batches of $\x_0$ are shown on the left. They are quite arbitrary and clearly non-Gaussian. For reference, the corresponding distributions of $\x_1$ are shown on the right -- they are close to Gaussian as expected. If the encoder $\E_\Delta$ can produce unit Gaussian variable $\z$ under such kind of arbitrarily distributed $\x_0$, when the distributions of $\x_0$ are arbitrarily different across different batches and arbitrarily changed at the beginning of each training iteration (i.e., epoch), it is reasonable to state that the output unit Gaussian $\z$ shall be independent of $\x_0$.

\subsubsection{Statistical Distance} \label{sec:statdis}

There exist a variety of statistical distances that one can use in defining $\mathcal{L}_{\textrm{distance}}$ in the distributional loss \eqref{LossD}. Through numerical experimentation, we have found Renyi-entropy-based distance (\cite{anderson1994two}) to be highly effective.
The Renyi-entropy-based distance first estimates the probability density distribution of the samples using kernel density estimator (KDE) (c.f. \cite{dehnad1987density}) and then computes the $L^2$ distance between the KDE estimated distribution to the target distribution. In our case, consider a batch $ B = \left\{\z^{(i)}, i=1,\dots, N\right\} $ with $N$ samples. Its KDE  distribution estimation is
$$
    \hat{f}_B(\mathbf{y})=\frac{1}{N} \sum_{i=1}^N \kappa_\sigma\left(\mathbf{z}^{(i)}-\mathbf{y}\right),
    $$
    where $\kappa_\sigma$ is a kernel function with bandwidth $\sigma>0$.
The Renyi-entropy-based distance loss is defined as 
\be 
   \mathcal{L}_{\textrm{distance}}(B) = \norm{\hat{f}_B - {f}_{\mathcal{N}}}_2,
    \label{kde:l2loss}
\ee
where $f_{\mathcal{N}}$ is the probability distribution function of $n_z$-dimensional unit Gaussian.

Mathematically speaking, minimization of this loss function to zero value shall enforce the distribution of the batch $B$ to follow the unit Gaussian. In practice, however, this can not be accomplished, due to the discrete nature of the batch $B$ and its finite number of samples, along with the limited capability of (stochastic) optimization method used during the training. (No optimization method can achieve zero loss value.) Our numerical experimentation suggests it is necessary to incorporate a measure of the statistical moments to enhance the learning.
    


\subsubsection{Moment Loss Function} \label{sec:moment}

To enhance the performance of the training and further enforce the batch $B$ to follow unit Gaussian distribution, we introduce moment loss $\mathcal{L}_{\textrm{moment}}(B)$ in the distributional loss \eqref{LossD}. More specifically, we seek to enforce the marginal moments of the batch $ B = \left\{\z^{(i)}, i=1,\dots, N\right\}$ to match those of the unit Gaussian, for up to the 6th central moment.

We first consider the case of $n_z=1$. The moment loss function takes the form
\be \label{1dmoment}
    \mathcal{L}_{\textrm{moment}}(B) = \sum_{j=1}^{J}\frac{1}{c_j}\left[\widehat{\mu}^{(j)}(B)-\mu^{(j)}(P_\mathcal{N})\right]^2,
\ee
wherefor each $1\leq j\leq J$, $\widehat{\mu}^{(j)}(B)$ is the $j$-th central moment of the batch $B$, $\mu^{(j)}(P_\mathcal{N})$ the $j$-th central moment of unit Gaussian, $c_j>0$ a scaling parameter, and $J$ the highest moment chosen.   We set $J=6$ in all of our numerical testing and found it to be a robust choice.

For unit Gaussian, we have $\mu^{(1)}=\mu^{(3)}=\mu^{(5)}=0$, $\mu^{(2)}=1$, $\mu^{(4)}=3$, and $\mu^{(6)}=15$. After extensive testing, we advocate the use of the following scaling parameters:
\be \label{moment_para}
c_1 = 1.0, \quad c_2=1.0, \quad c_3=2.0, \quad c_4=3.0, \quad c_5=8.0, \quad c_6=15.0.
\ee

For multi-dimensional case $n_z>1$, the moments are written as $n_z$-dimensional vectors consisting of the marginal moments in each direction, and vector 2-norm is used in \eqref{1dmoment}. We also introduce cross-correlation coefficients among all pairs of dimensions, which are zero for the unit Gaussian distribution. The moment loss for $n_z>1$ is thus defined as
\be \label{ndmoment}
    \mathcal{L}_{\textrm{moment}}(B) = \sum_{j=1}^{6}\frac{1}{c_j}\left\|\widehat{\mu}_N^{(j)}(B)-\mu^{(j)}(P_\mathcal{N})\right\|_2^2 + \frac{\nu}{K} \sum_{1\leq j<k \leq n_z}\left(\widehat{\rho}_{jk}(B)\right)^2,
\ee
where $\widehat{\rho}_{jk}(B)$, $1\leq j<k\leq n_z$, are the cross-correlation coefficient of the batch $B$, $K=n_z(n_z-1)/2$ the total number of the cross-correlation coefficients, and $\nu>0$ a penalty parameter. Upon extensive numerical experimentation, we suggest the use of $\nu=2$.
\section{Numerical Examples}
\label{sec:examples}

In this section, we present numerical tests to demonstrate the performance of the proposed autoencoder sFML method. The examples cover the following representative cases:
\begin{itemize}
\item Linear SDEs: Ornstein-Uhlenbeck (OU) process and geometric Brownian motion;
\item Nonlinear SDEs: SDEs with exponential and trigonometric drift or diffusion, and SDE double-well potential;
\item SDEs with non-Gaussian noise of exponential and lognormal distributions;
\item Multi-dimensional SDEs: 2-dimensional and 5-dimensional OU processes. These examples demonstrate how the training procedure can automatically detect the correct dimension $n_z$ for the latent random variable.
\end{itemize}

For all these problems, the true SDEs are used to generate training data sets, in the form of \eqref{data_set}, as well as validation data to examine the accuracy of the prediction results by the learned sFML model. The knowledge of the true SDEs is otherwise not used anywhere in the model construction procedure. 
The trajectory data \eqref{traj} are generated by solving the SDEs via Euler-Maruyama method, with \(N_T=10,000\) initial conditions uniformly distributed within a region (to be specified for each example) and a length $L=100$ with a time step \(\Delta = 0.01\). These \(N_T\) trajectories thus cover a time up to $T=1.0$. They are subsequently reorganized via the procedure in Section \ref{sec:math}, resulting in training data sets \eqref{data_set} with $M=10^6$ data pairs. In all of the examples, we re-sample the training data sets into  \(n_{B} = 1,000\) batches, each of which contains \(N=10,000\) data pairs, by following the sub-sampling strategy from Section \ref{sec:resample}.

Regarding the DNN architecture, we employ fully connected feedforward DNNs for both the encoder and decoder. The encoder has  4 layers, each of which with 20 nodes. The first three layers utilize the eLU activation function.  The decoder has the same structure as the encoder, except an identity operation is introduced to implement it in the ResNet structure. All the examples underwent training for up to 1,000 epochs. 
 
To evaluate the efficacy of the method, we conduct system predictions using the learned sFML models for a time horizon typically up to $T=5\sim 500$, much longer than that of the training data (whose time horizon is up to $T=1$.) We then compare the sFML prediction results against the ground truth obtained by solving the true SDEs. The comparisons include the following:
\begin{itemize}
    \item Mean and standard deviation of the solution;
    \item Evolution of the solution probability distribution  over time;  
    \item The one-step conditional distribution $\mathbb{P}(\x_{n+1}|\x_{n}=\x)$ for some arbitrarily chosen $\x$; 
    \item The effective drift and diffusion functions reconstructed from the learned sFML models against the true drift and diffusion;
    \item The distribution of the latent variables obtained from the encoder against the true latent variables (unit Gaussian).    
\end{itemize}

\subsection{Linear SDEs}

We first present the results of learning an Ornstein Uhlenbeck (OU) process and a geometric Brownian motion.

\subsubsection{OU Process}
The true SDE takes the following form,
\begin{equation} \label{eq: OUP}
    d x_t = \theta(\mu - x_t) d t + \sigma d W_t,
\end{equation}
where the parameters are set as $\theta=1.0$, $\mu=1.2$, and $\sigma=0.3$.

For the training data, the initial points are sampled uniformly from interval $(0,2.5)$. For the sFML model prediction, we fix the initial condition to be \(x_0=1.5\) and generate $500,000$ prediction trajectories up to \(T=5\) to examine the solution statistics and distribution.

\begin{figure}[htbp]
  \centering
  \includegraphics[width=.8\textwidth]{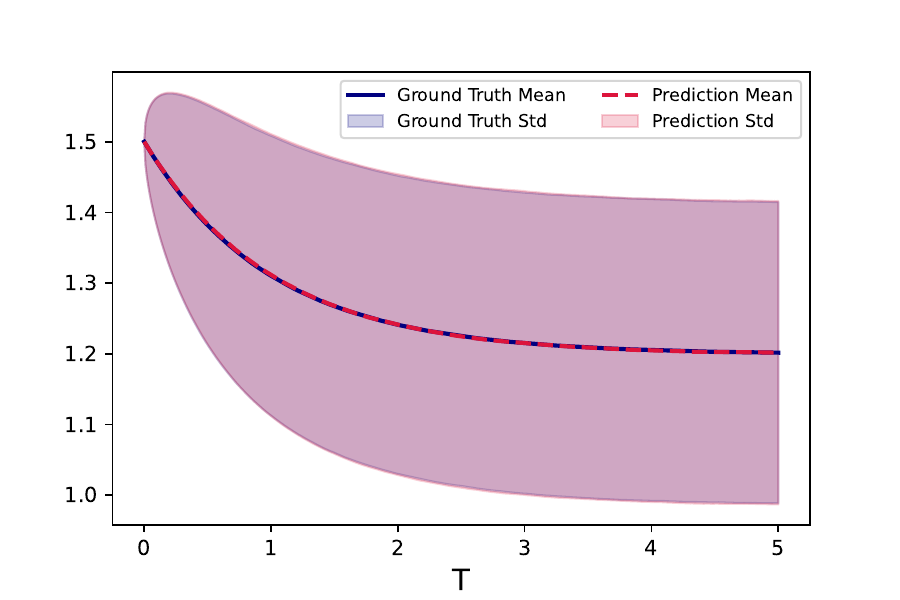}
  \caption{OU process \eqref{eq: OUP}: mean and standard deviation by the learned sFML model against the ground truth.}
    \label{fig:OU_stat}
\end{figure}

The mean and standard deviation of the predictions from the learned sFML model, along with the reference mean and standard deviation from the true SDE, are shown in Figure \ref{fig:OU_stat}. Good agreement between the learned sFML model and the true model can be observed. Note that the agreement goes beyond the time horizon of the training data by a factor of $5$.

\begin{figure}[htbp]
  \centering
  \includegraphics[width=.43\textwidth]{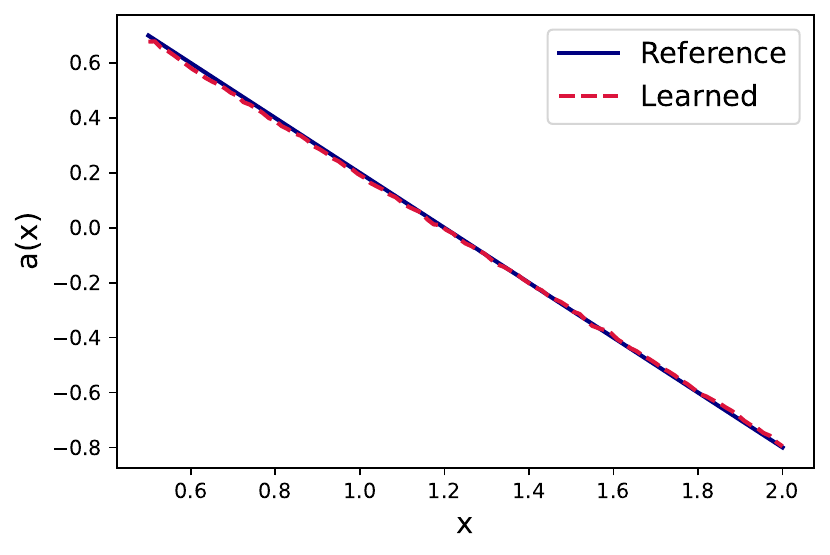}
  \includegraphics[width=.43\textwidth]{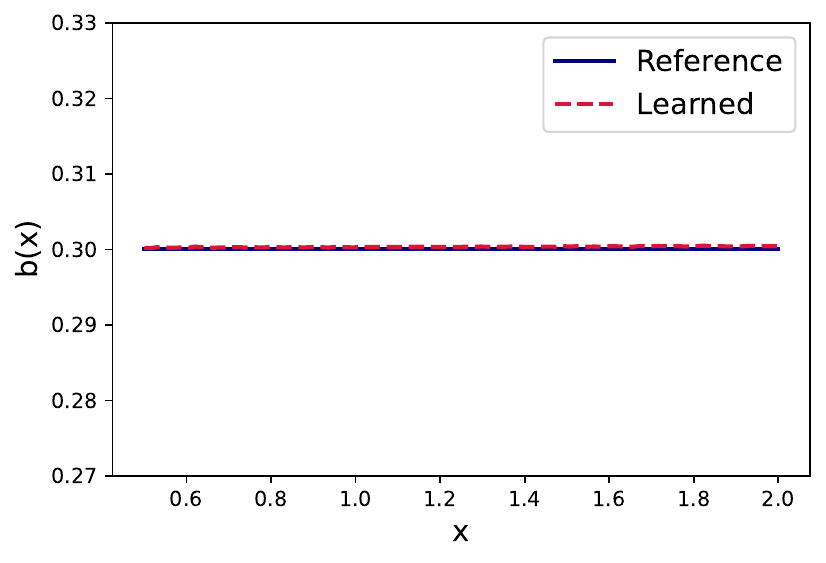}

  \caption{OU process \eqref{eq: OUP}: Learned and reference effective drift and diffusion of the sFML model. Left: drift $a(x)=1.2-x$; Right: diffusion $b(x)=0.3$.}
    \label{fig:OU_ab}
\end{figure}

We then recover the effective drift and diffusion of the sFML model. From Figure \ref{fig:OU_ab}, we observe they closely approximate the reference true functions with relative errors of \(O(10^{-3})\).

\begin{figure}[htbp]
  \centering
    \includegraphics[width=.43\textwidth]{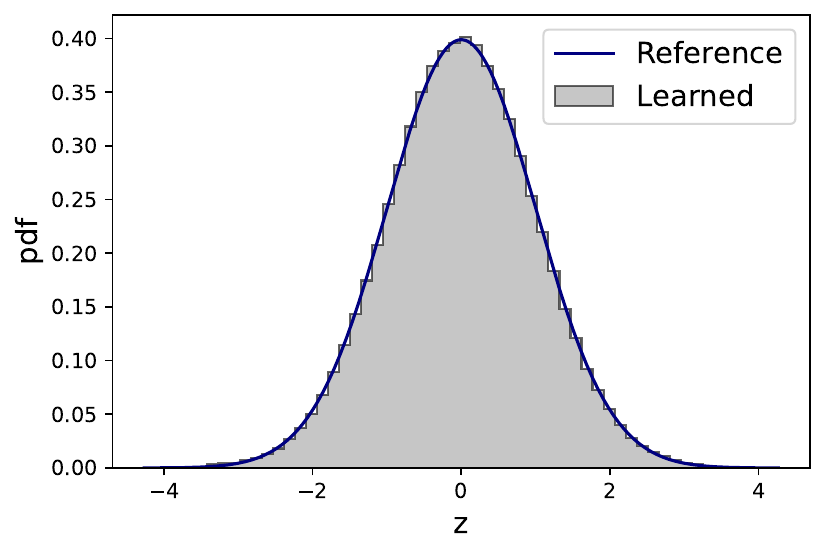}
    \includegraphics[width=.43\textwidth]{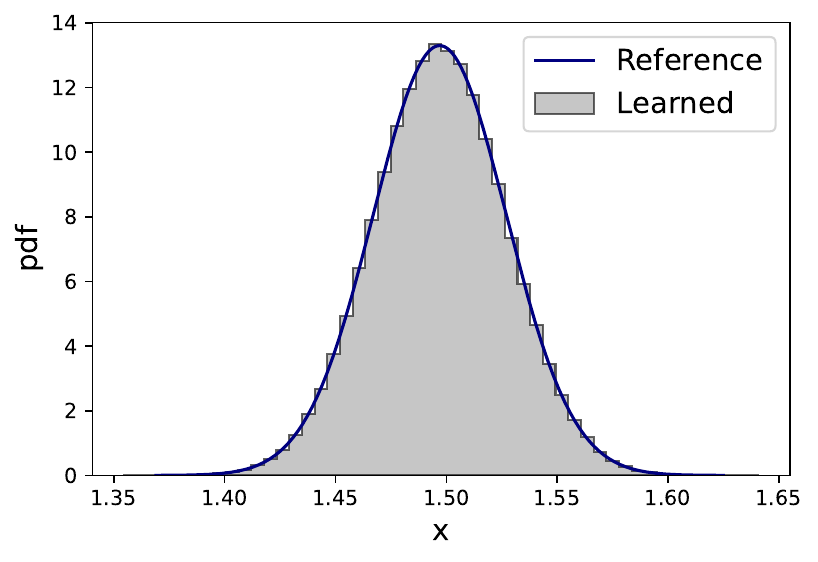}
  \caption{OU process \eqref{eq: OUP}: Left: Distribution of the latent variable from the trained encoder $\E_\Delta(x_0,x_1)$, $x_0 = 1.5$, against unit Gaussian $\mathcal{N}(0,1)$; Right: one-step conditional distribution of the trained decoder $\D_\Delta(1.5,z)$, $z\sim \mathcal{N}(0,1)$, against the ground truth.}
      \label{fig:OU_pdf}
\end{figure}

In Figure \ref{fig:OU_pdf}, we present two histograms to compare the solution distributions further. On the left, we show the comparison between the distribution of the latent variable against the true latent variable (unit Gaussian). 
On the right of the figure, we show the histogram of the decoder \(\D_\Delta(1.5,z)\) with \(z \sim N(0,1)\). This represents the one-step conditional distribution $\mathbb{P}(\x_{n+1}|\x_{n}=1.5)$. We observe a good agreement with the ground truth.

\subsubsection{Geometric Brownian Motion}

We now consider a geometric Brownian motion (GBM)
\begin{equation}  \label{eq: GBM}
d x_t = \mu x_t d t + \sigma x_t d W_t,
\end{equation}
where the parameters are set as  $\mu=2$ and \(\sigma=1\).

For the training data, the initial points are uniformly drawn from the interval \((0,2)\). For model prediction,  we fix the initial condition at \(x_0=0.5\) and produce \(500,000\) prediction trajectories to examine the solution statistics. Due to the exponential growth of this particular geometric Brownian motion over time, the predictive simulations are
stopped at \(T = 1.0\).

\begin{figure}[htbp]
  \centering
  \includegraphics[width=.8\textwidth]{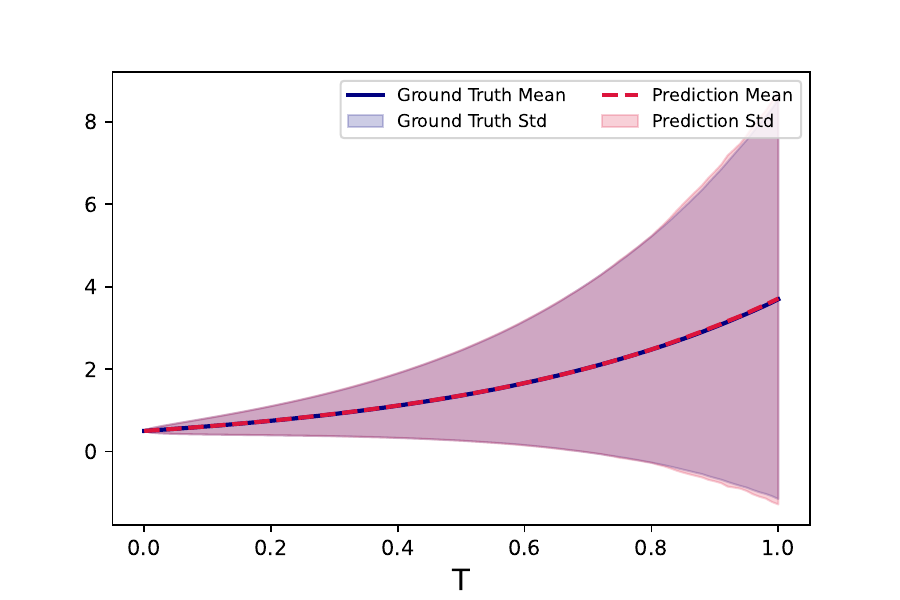}
  \caption{Geometric Brownian Motion \eqref{eq: GBM}: mean and standard deviation by the sFML model.}
    \label{fig:GB_stat}
\end{figure}

\begin{figure}[htbp]
  \centering
  \includegraphics[width=.43\textwidth]{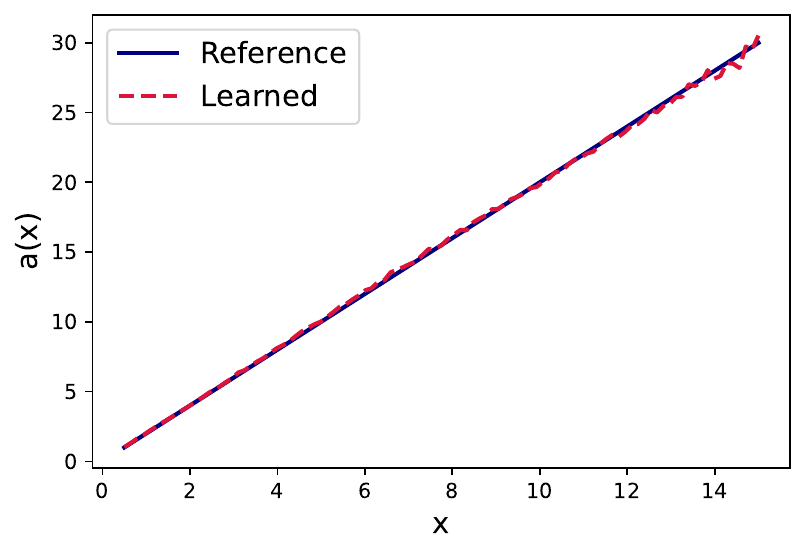}
  \includegraphics[width=.43\textwidth]{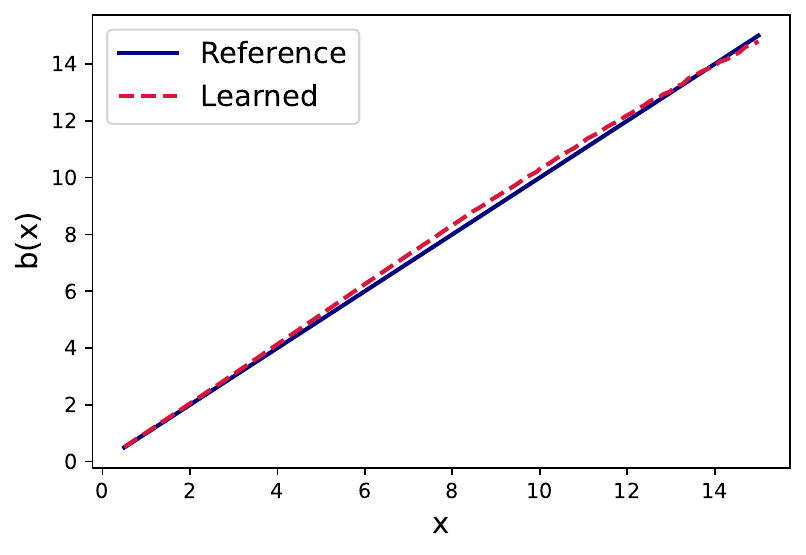}

  \caption{Geometric Brownian Motion \eqref{eq: GBM}: Learned and reference effective drift and diffusion of the sFML model. Left: drift $a(x)=2x$; Right: diffusion $b(x)=x$.}
    \label{fig:GB_ab}
\end{figure}

\begin{figure}[htbp]
  \centering
    \includegraphics[width=.43\textwidth]{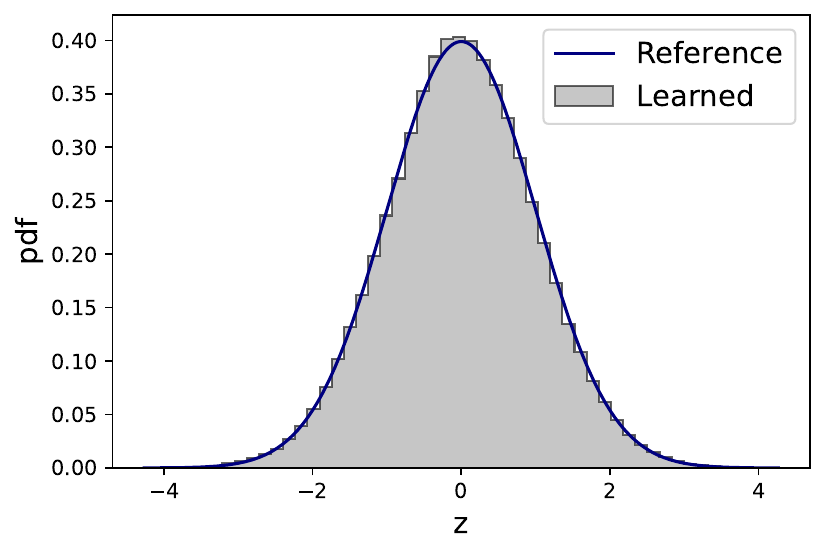}
    \includegraphics[width=.43\textwidth]{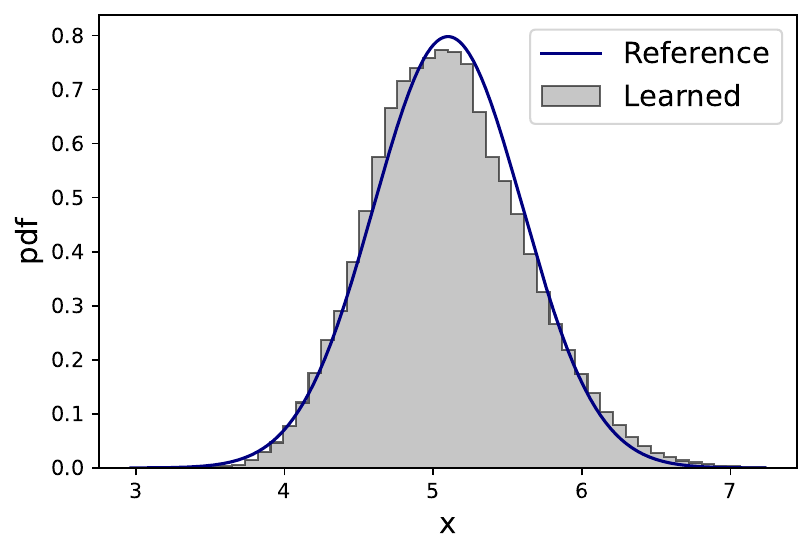}
  \caption{Geometric Brownian Motion \eqref{eq: GBM}: Left: distribution of the latent variable from the trained encoder $\E_\Delta(x_0,x_1),\, x_0 = 5.0$ against unit Gaussian; Right: one-step conditional distribution of the trained decoder $\D_\Delta(5.0,z),\, z\sim \mathcal{N}(0,1)$, against the ground truth.}
     \label{fig:GB_pdf}
\end{figure}

The mean and standard deviation of predictions are shown in Figure \ref{fig:GB_stat}. The recovered effective drift and diffusion functions by the sFML model are presented in Figure \ref{fig:GB_ab}. To further validate the learned sFML model, we compare the outputs of the encoder and decoder against their corresponding references in terms of distributions, as displayed in Figure \ref{fig:GB_pdf}. Good agreement with the ground truth can be observed throughout these results.

\subsection{Nonlinear SDEs}

We now consider two It\^{o} type SDEs with non-linear drift and diffusion functions, along with an SDE with a double well potential for long-term predictions.

\subsubsection{SDE with nonlinear diffusion}
We first consider an SDE with a nonlinear diffusion,
\begin{equation} \label{eq: exp}
    d x_t = -\mu x_t d t+\sigma e^{-x_t^2} d W_{t},
\end{equation}
where $\mu$ and $\sigma$ are constants. In this example, we set $\mu=5$ and $\sigma=0.5$.

\begin{figure}[htbp]
  \centering
  \includegraphics[width=.8\textwidth]{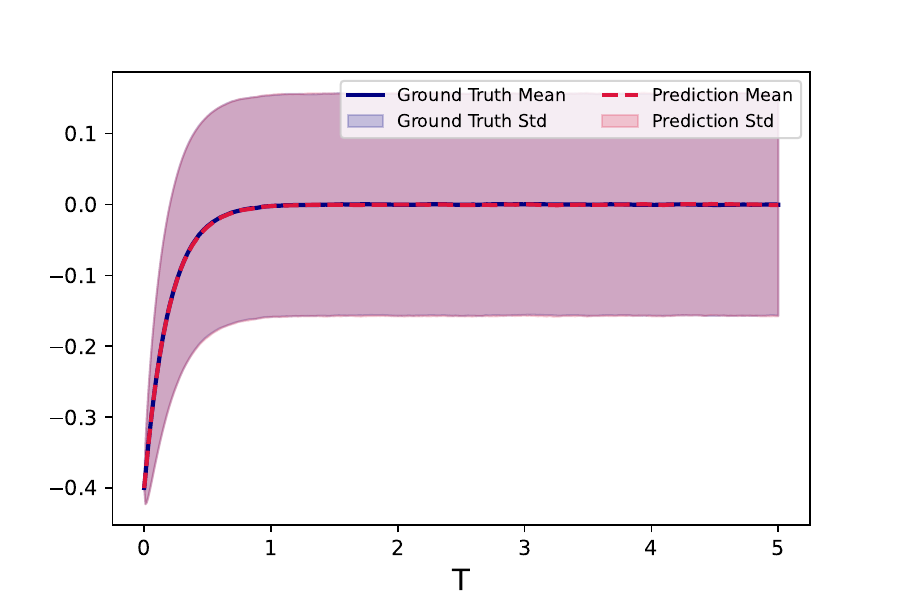}
  \caption{SDE with nonlinear diffusion \eqref{eq: exp}: mean and standard deviation of the sFML model.}
  \label{fig:Exp_stat}
\end{figure}

\begin{figure}[htbp]
  \centering 
  \includegraphics[width=.42\textwidth]{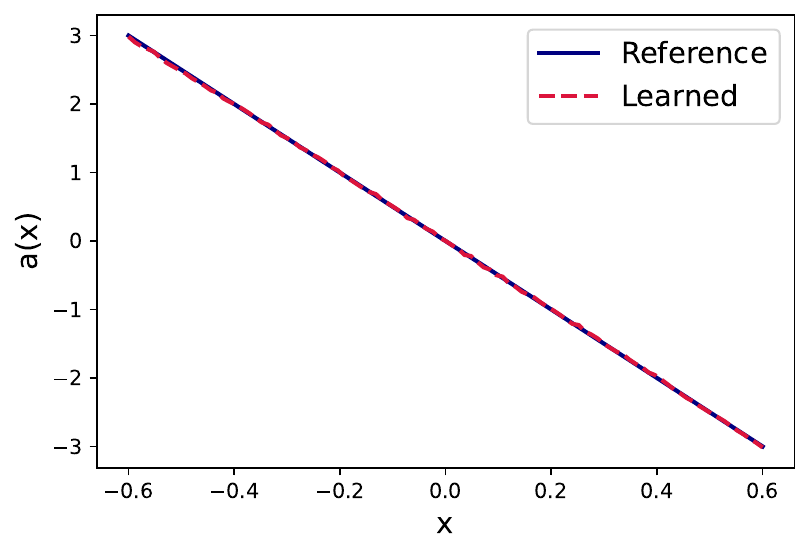}
  \includegraphics[width=.43\textwidth]{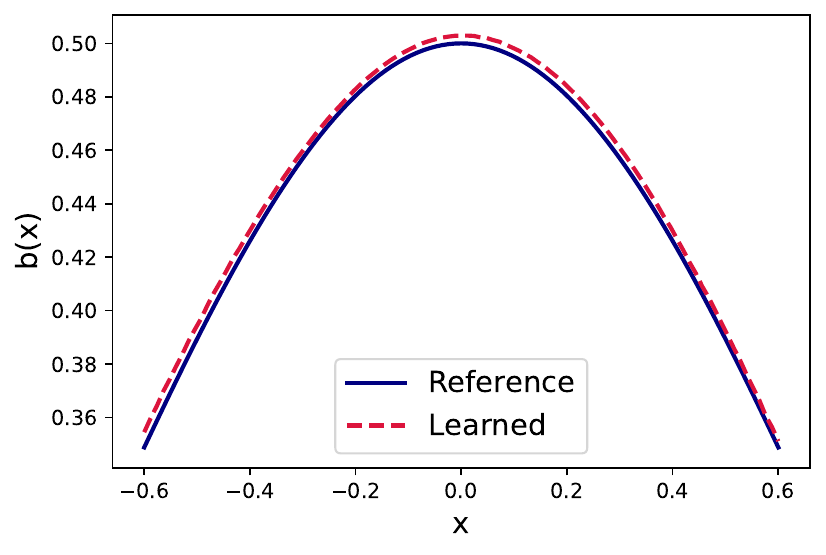}
  \caption{SDE with nonlinear diffusion \eqref{eq: exp}: Learned and reference effective drift and diffusion of the sFML model. Left: drift $a(x)=-5x$; Right: diffusion $b(x)=0.5e^{-x^2}$.}
   \label{fig:Exp_ab}
\end{figure}

\begin{figure}[htbp]
  \centering
    \includegraphics[width=.43\textwidth]{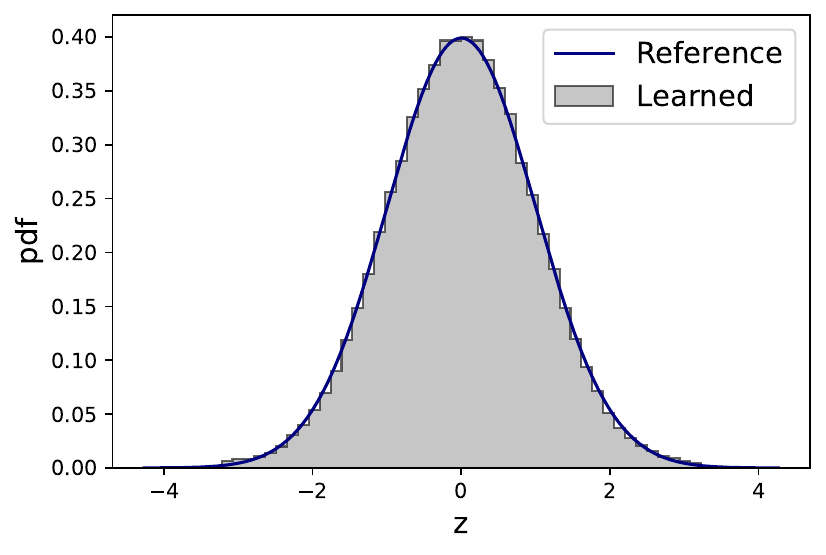}
    \includegraphics[width=.415\textwidth]{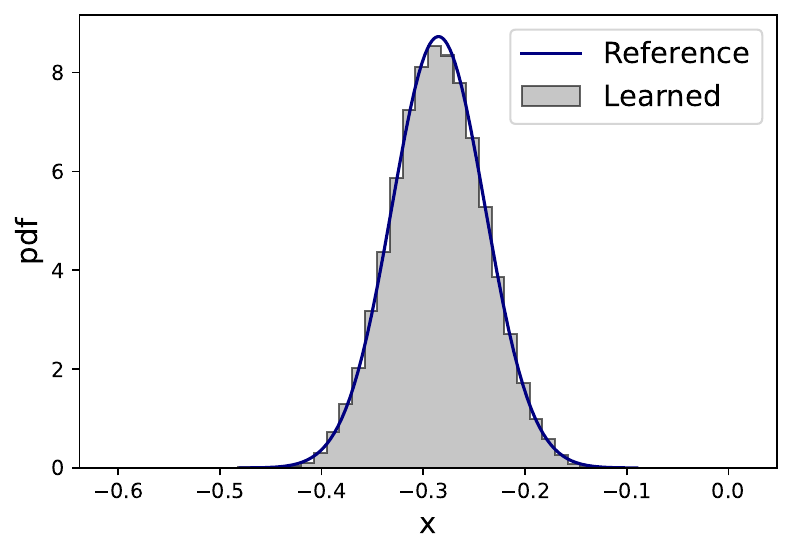}
  \caption{SDE with nonlinear diffusion \eqref{eq: exp}: Left: distribution of the latent variable from the trained encoder $\E_\Delta(x_0,x_1),\, x_0 = -0.3$; Right: one-step conditional distribution of the trained decoder $\D_\Delta(-0.3,z),\, z\sim \mathcal{N}(0,1)$, against the ground truth.}
  \label{fig:Exp_pdf}
\end{figure}

For the training data, the SDE is solved with initial conditions drawn uniformly from $(-1,1)$. Upon constructing the sFML model, we conduct system prediction up to $T=5.0$. The evolution of mean and STD of the learned sFML model, with a fixed initial condition $x_0 = -0.4$,  are shown in Figure \ref{fig:Exp_stat}. The recovered effective drift and diffusion functions from the learned sFML model are plotted in Figure \ref{fig:Exp_ab}, and the comparisons between the encoder and decoder and their respective references in terms of distributions can be found in Figure \ref{fig:Exp_pdf}. Again, good agreements can be observed between the learned sFML model and the ground truth.

\subsubsection{Trigonometric SDE}

We now consider the following non-linear SDE with trigonometric drift and diffusion,
\begin{equation} \label{eq: Trig}
    d x_t = \sin(2k\pi x_t) d t+\sigma \cos(2k\pi x_t) d W_{t},
\end{equation}
where the constants are set at $k=1$ and $\sigma=0.5$.

\begin{figure}[htbp]
  \centering
  \includegraphics[width=.8\textwidth]{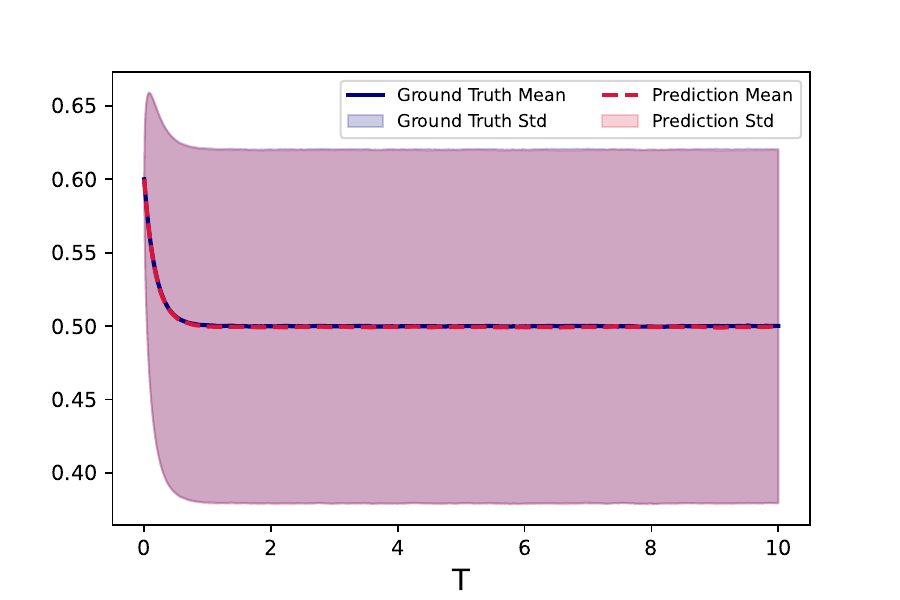}
  \caption{Trigonometric SDE \eqref{eq: Trig}: Mean and standard deviation of the sFML model. } 
  \label{fig:Trig_stat}
\end{figure}

\begin{figure}[htbp]
  \centering
    \includegraphics[width=.43\textwidth]{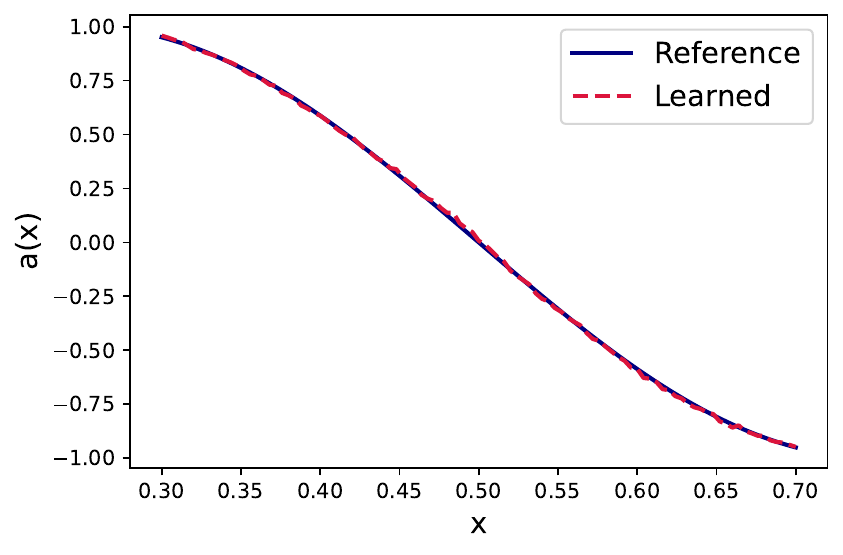}
  \includegraphics[width=.43\textwidth]{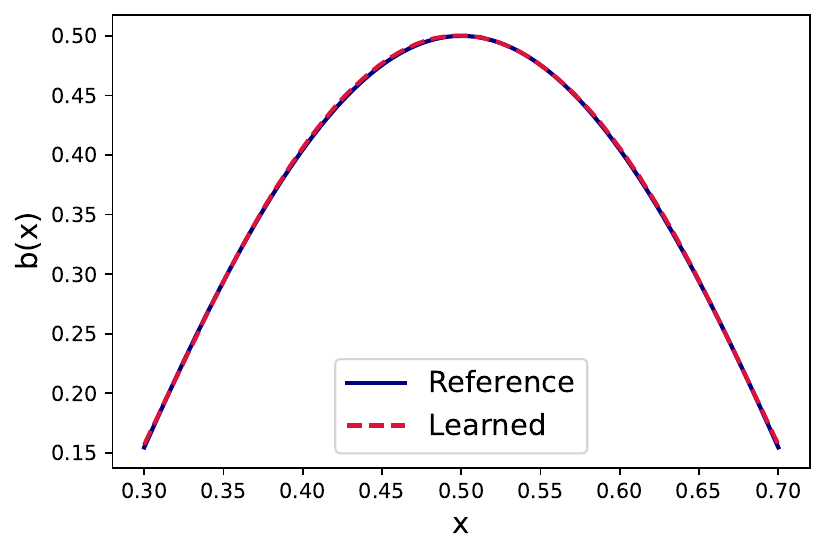} 
  \caption{Trigonometric SDE \eqref{eq: Trig}: Learned and reference effective drift and diffusion of the sFML model. Left: drift $a(x)=\sin(2k\pi x)$; Right: diffusion $b(x)=\cos(2k\pi x)$.}
  \label{fig:Trig_ab}
\end{figure}

\begin{figure}[htbp]
  \centering
    \includegraphics[width=.43\textwidth]{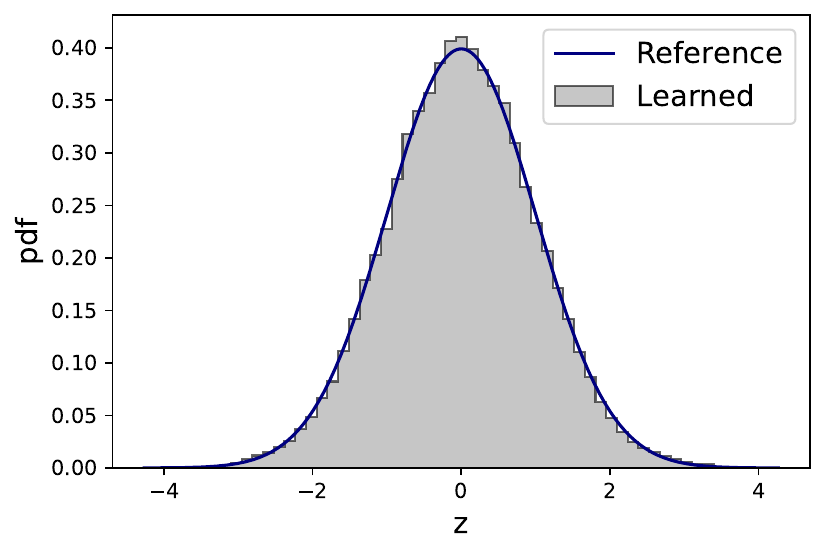}
    \includegraphics[width=.42\textwidth]{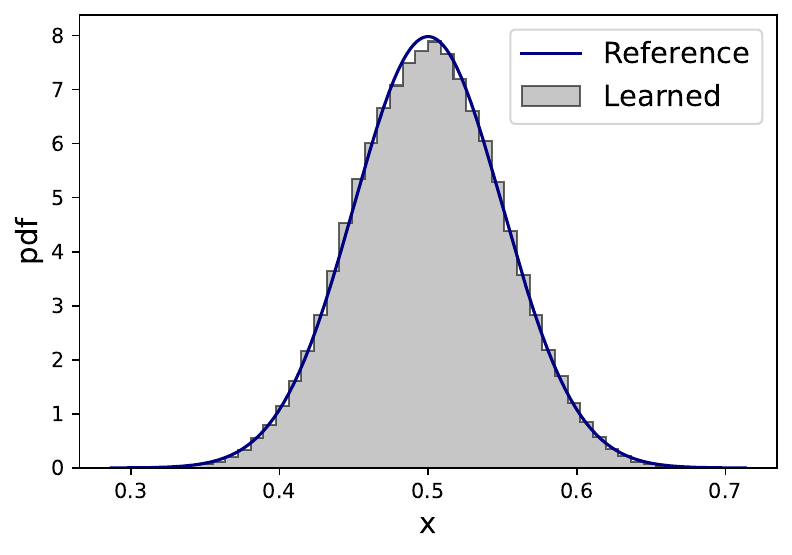}
  \caption{Trigonometric SDE \eqref{eq: Trig}: Left: distribution of the latent variable of the trained encoder $\E_\Delta(x_0,x_1),\, x_0 = 0.5$ against unit Gaussian; Right: one-step conditional distribution of the trained decoder $\D_\Delta(0.5,z),\, z\sim \mathcal{N}(0,1)$, against the ground truth.}
  \label{fig:Trig_pdf}
\end{figure}

The training data are generated with initial conditions uniformly
distributed as $(0.35,0.7)$. Upon training the sFML model, we carry out the predictions with an initial condition $x_0 = 0.6$ 
for time up to $T = 10.0$.  The mean and standard deviation of the predictions from the learned sFML model are shown in Figure \ref{fig:Trig_stat}. As illustrated by Figure \ref{fig:Trig_ab}, the recovered effective drift and diffusion functions from the sFML model compare very well with the true drift and diffusion functions. From Figure \ref{fig:Trig_pdf}, we also observe excellent agreement between the distributions from the encoder and decoder and the reference solutions. Note that the same example was considered in \cite{chen2023learning}, where accuracy deterioration was observed over the prediction time. Here, we observe that the high accuracy of the sFML model maintains in a more robust manner over the prediction time.

\subsubsection{SDE with Double Well Potential}

We consider an SDE with a double well potential:
\begin{equation} \label{eq: doublewell}
    d x_t = (x_t-x_t^3) d t+\sigma d W_{t},
\end{equation}
where the constant is set as $\sigma=0.5$. The stochastic driving term is sufficiently large so that the solution has random transitions between the two stable states $x=1$ and $x=-1$.

The training data are generated by solving the true SDE with initial conditions uniformly sampled in $(-2.5,2.5)$  up to $T=1.0$.

\begin{figure}[htbp]
  \centering
  \includegraphics[width=.8\textwidth]{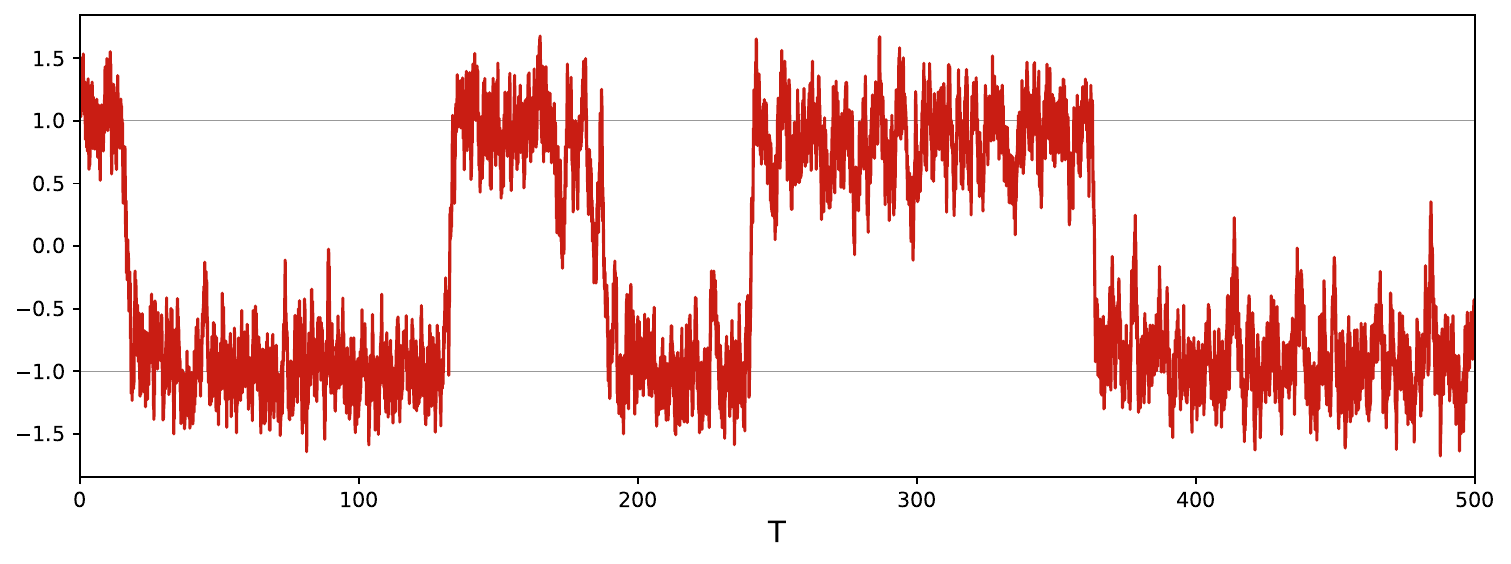}
  \caption{SDE with Double Well Potential \eqref{eq: doublewell}: Solution trajectory for time up to $T=500$ from initial condition $x_0 = 1.5$, simulated by the learned sFML model. The random switching between the two stable states $x=\pm 1$ is clearly visible.}
   \label{fig:DW_sample}
\end{figure}

\begin{figure}[htbp]
  \centering
  \includegraphics[width=.24\textwidth]{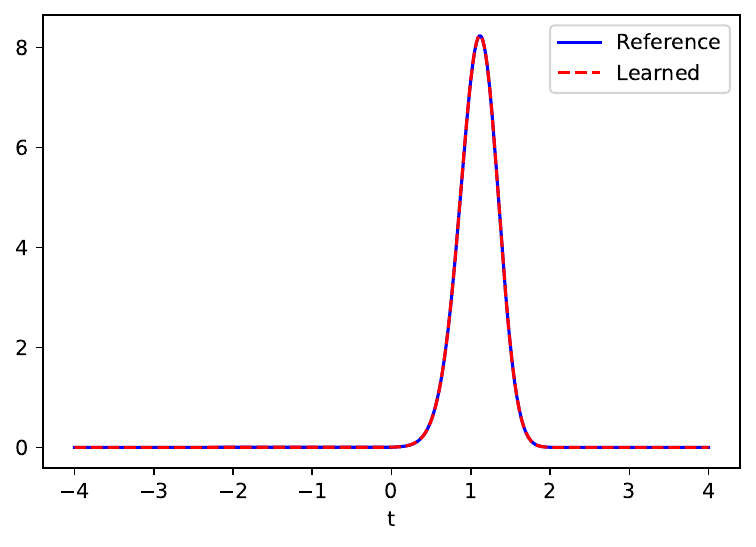}
  \includegraphics[width=.24\textwidth]{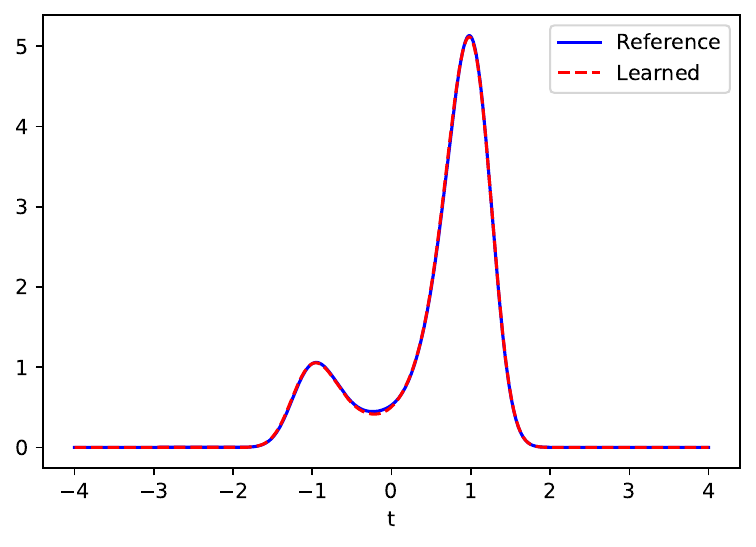}
  \includegraphics[width=.24\textwidth]{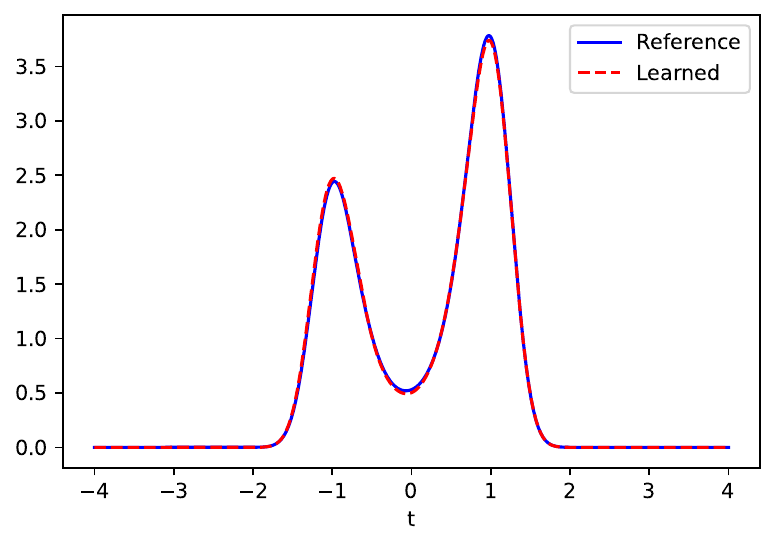}
  \includegraphics[width=.24\textwidth]{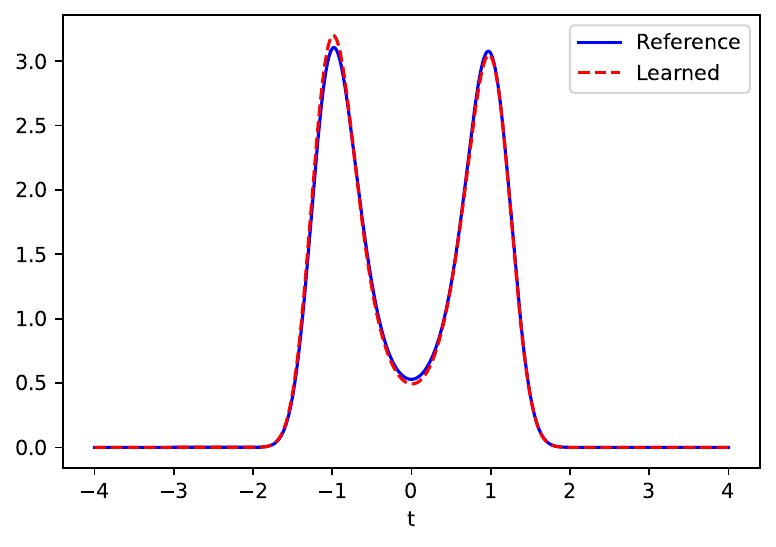}
  \caption{SDE with Double Well Potential \eqref{eq: doublewell}: Evolution of the solution PDF of the learned sFML model with an initial condition $x_0=1.5$, at time $T=0.5$, $10$, $30$, and $100$ (from left to right).}
      \label{fig:DW_pdf}
\end{figure}

\begin{figure}[htbp]
  \centering
  \includegraphics[width=.42\textwidth]{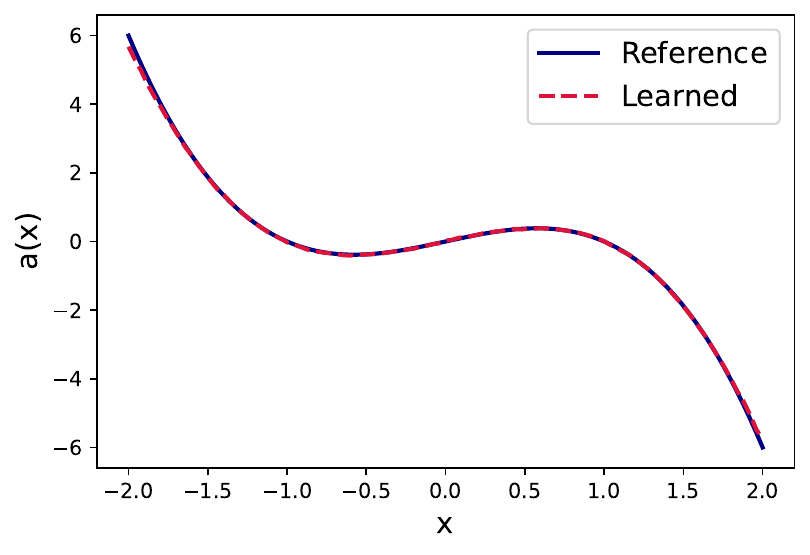}
  \includegraphics[width=.43\textwidth]{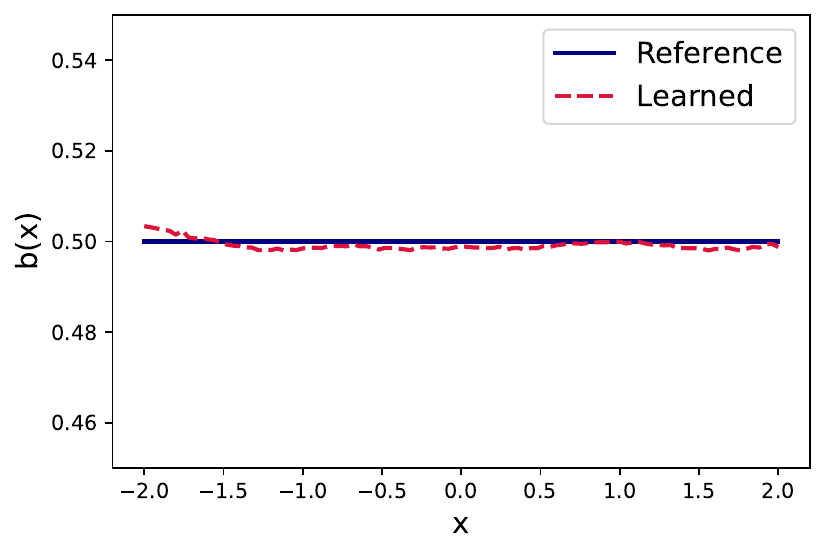}
  \caption{SDE with Double Well Potential \eqref{eq: doublewell}: Learned and reference effective drift and diffusion of the sFML model. Left: recovery of the drift $a(x)=x-x^3$; Right: recovery of diffusion $b(x)=0.5$.}
   \label{fig:DW_show}
\end{figure}

Upon constructing the sFML model, we conduct system prediction for up to $T=500.0$. This is a long-term simulation well beyond the time window of the training data ($T=1.0$). One sample prediction trajectory with an initial condition $x_0=1.5$ is shown in Figure \ref{fig:DW_sample}, where we clearly observe the random switching between the two stable states $x=\pm 1$. The switching time is on the order of $O(100)$ and well outside the training data time window ($T=1$). Due to the random transitions between the two stable states, the probability distribution of the solution becomes bi-modal over time and reaches a stable state asymptotically. To verify this, we conduct ensemble sFML predictions (from the same initial condition $x_0=1.5$) of $500,000$ sample realizations to collect the solution statistics.  The probability distribution of the solutions by the sFML model can be seen from Figure \ref{fig:DW_pdf}, at time levels $T = 0.5, 10.0, 30.0, 100.0$. We observe excellent agreement between the sFML model predictions and the ground truth, for such a long-term simulation. We also recover the effective drift and diffusion functions of the sFML model and observe their good agreement with the ground truth in Figure \ref{fig:DW_show}.

\subsection{SDEs with Non-Gaussian Noise}

We now present the proposed sFML approach for modeling SDEs driven by non-Gaussian stochastic processes.

\subsubsection{Noise with Exponential Distribution}

We consider an SDE with exponentially distributed noise,
\begin{equation}\label{eq: EXPN}
d x_t=\mu x_t d t+\sigma \sqrt{d t} ~\eta_t, \qquad \eta_t \sim \operatorname{Exp}(1),
\end{equation}
where $\eta_t$ has an exponential pdf $f_\eta(x)=e^{-x}$, $x \geq 0$, and the constants are set as $\mu=-2.0$ and $\sigma=0.1$.
 
\begin{figure}[htbp]
  \centering
  \includegraphics[width=.49\textwidth]{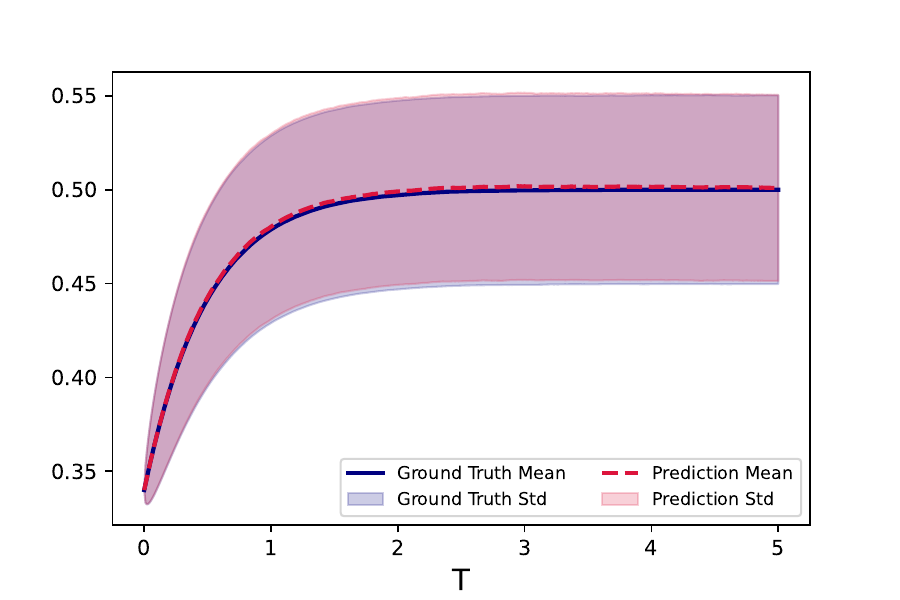}
    \includegraphics[width=.44\textwidth]{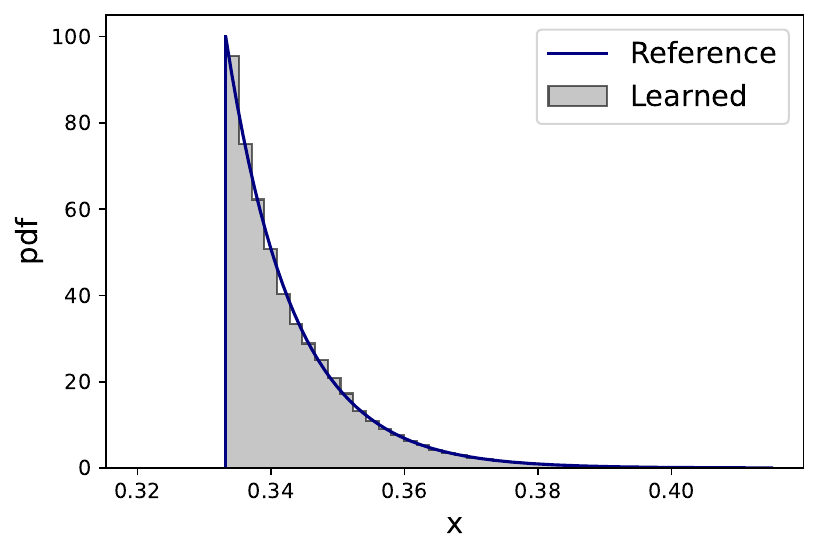}
  \caption{SDE with exponentially distributed noise \eqref{eq: EXPN}: Left: mean and standard
    deviation of the sFML model predictions; Right: one-step conditional probability by the trained decoder $\D_\Delta(x,z),\, z\sim \mathcal{N}(0,1)$,  against that of the true model at $x=0.34$.}
      \label{fig:Expdis_pdf}
\end{figure}

The training data are generated by solving the true SDE with initial conditions uniformly sampled in $(0.2,0.9)$  for up to $T=1.0$. Upon constructing the sFML model, we report the simulation result with an initial condition  $x_0=0.34$ for up to $T=5.0$. The mean and standard deviation of the predictions are shown on the left of Figure \ref{fig:Expdis_pdf}. The one-step conditional distribution by the sFML model through the trained decoder $\D_\Delta(x,z)$, $z\sim \mathcal{N}(0,1)$, is shown for an arbitrarily chosen location at $x=0.34$, on the right of Figure \ref{fig:Expdis_pdf}. We observe excellent agreement between the learned sFML model prediction and the reference solution.

\begin{figure}[htbp]
  \centering
  \includegraphics[width=.43\textwidth]{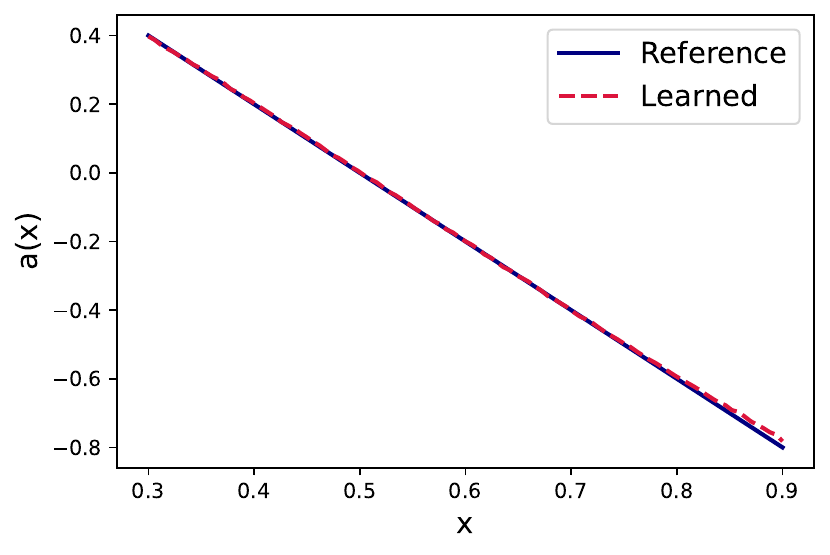}
  \includegraphics[width=.43\textwidth]{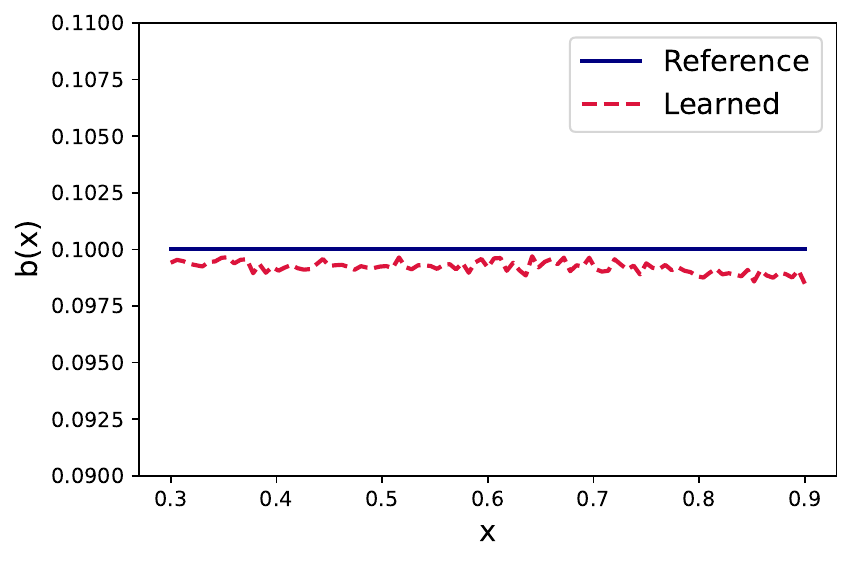}
  \caption{SDE with exponentially distributed noise \eqref{eq: EXPN}: Learned and reference effective drift and diffusion of the sFML model. Left: recovery of the drift $a(x)=\mu x + \sigma/\sqrt{\Delta}$; Right: recovery of the diffusion  $b(x)=\sigma$. See \eqref{ab_exponentialnoise} for the reference solution.}
  \label{fig:Expdis_show}
\end{figure}

The exact drift and diffusion can be computed from the SDE by extracting the non-zero mean of the exponential distribution out of the noise term. We obtain
\begin{equation}
\begin{aligned}
& a\left(x_n\right)=\mathbb{E}\left(\left.\frac{x_{n+1}-x_n}{\Delta} \right| x_n\right)=\mu x_n+\frac{\sigma}{\sqrt{\Delta}}, \\
& b\left(x_n\right)=\operatorname{Std}\left(\left.\frac{x_{n+1}-x_n}{\Delta} \right| x_n\right)=\sigma .
\end{aligned}
\label{ab_exponentialnoise}
\end{equation}
The effective drift and diffusion recovered from the sFML model are shown in Figure \ref{fig:Expdis_show}, where good agreement with the reference solution can be seen.

\subsubsection{SDE with lognormally distributed noise}

We now consider the following stochastic system
\begin{equation}\label{eq: log}
d \log x_t=\left(\log m-\theta \log x_t\right) d t+\sigma d W_t,
\end{equation}
where $m, \theta$ and $\sigma$ are parameters. This is effectively an OU process after taking an exponential operation. Its dynamics can be solved by using the Euler-Maruyama method with the following scheme:
\begin{equation}
x_{n+1}=m^{\Delta} x_n^{1-\theta \Delta} \eta_n^{\sigma \sqrt{\Delta}}, \qquad \eta_n \sim \operatorname{Lognormal}(0,1) .
\end{equation}

\begin{figure}[htbp]
  \centering
  
  \includegraphics[width=.49\textwidth]{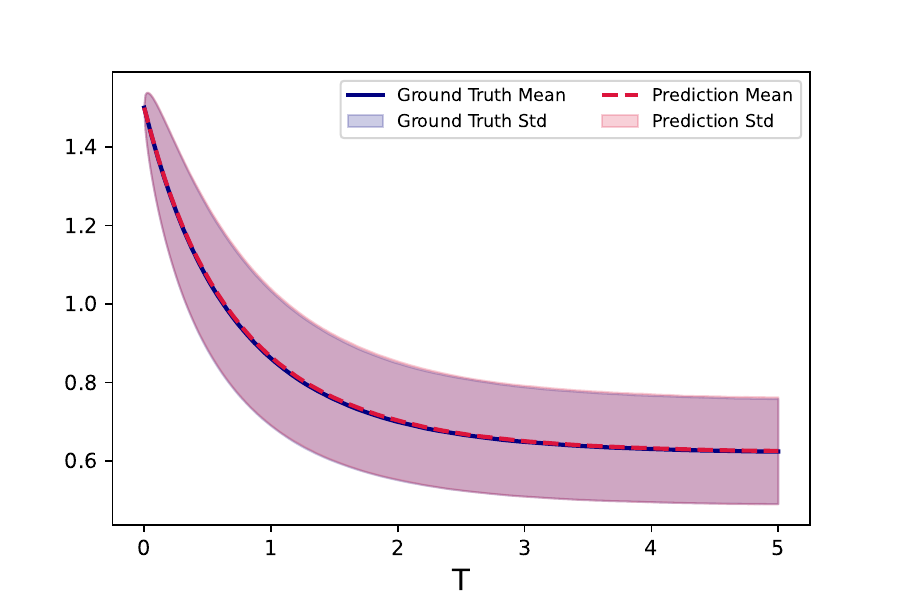}
  \includegraphics[width=.44\textwidth]{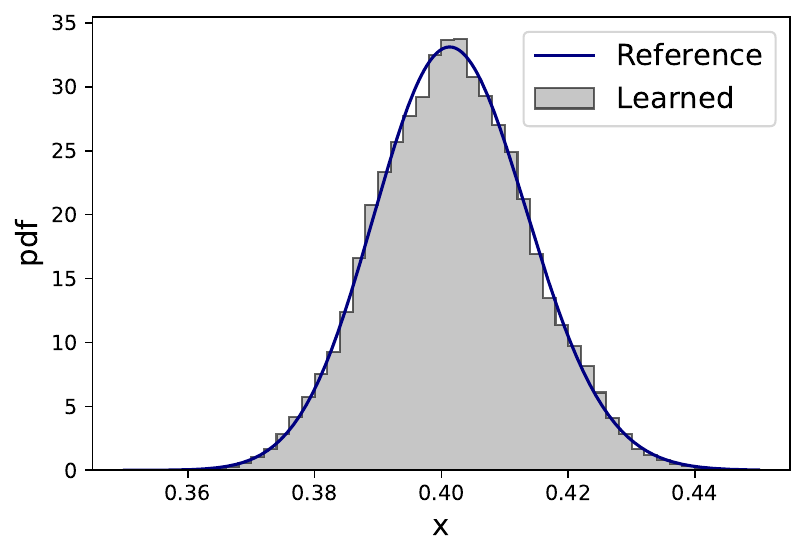}
  
  \caption{SDE with lognormally distributed noise \eqref{eq: log}. Left: mean and standard deviation of the sFML model prediction with the initial condition $x_0=1.5$; Right: one-step conditional probability by the trained decoder $\D_\Delta(0.5,z)$, $z\sim \mathcal{N}(0,1)$,  against the true model at $x=0.4$.}
    \label{fig:ExpOU_pdf}
\end{figure}

Here we set $m=1 / \sqrt{e}$, $\theta=1.0$, $\sigma=0.3$, and generate the training data with initial conditions uniformly sampled from $(0.2,0.9)$  for up to time $T = 1.0$. Upon
training the sFML model, we conduct system predictions with the sFML model for time up to $T = 5.0$. The mean and standard deviation from the predictions, as well as the one-step conditional distribution, are shown in Figure \ref{fig:ExpOU_pdf}. Good agreement with the reference solutions from the true SDE can be observed.

\begin{figure}[htbp]
  \centering
  
  \includegraphics[width=.43\textwidth]{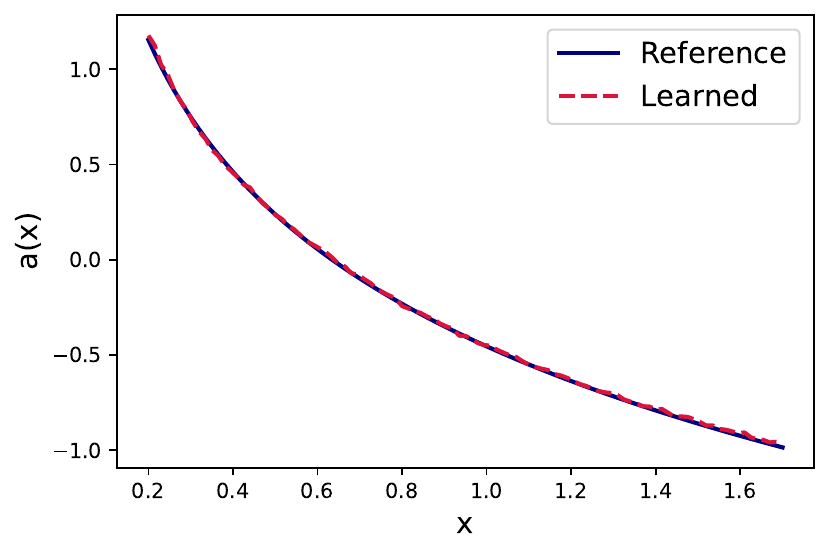}
  \includegraphics[width=.43\textwidth]{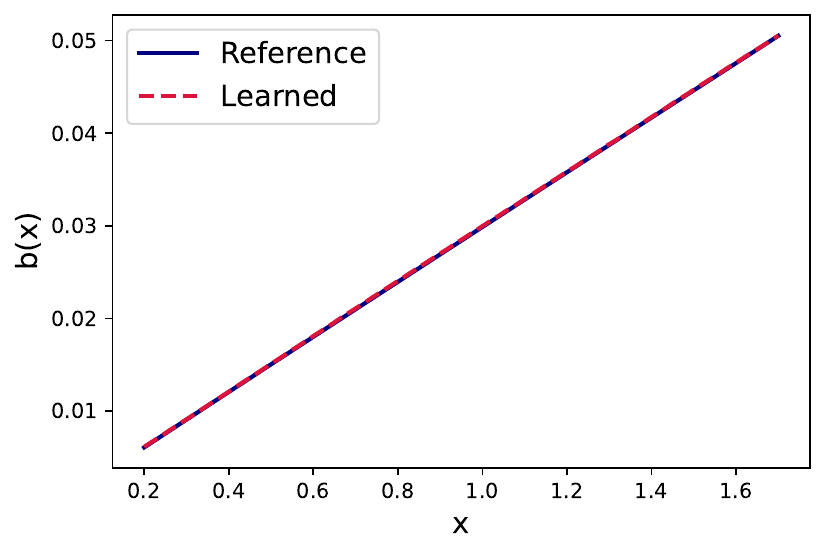}
    
  \caption{SDE with lognormally distributed noise \eqref{eq: log}: Learned and reference effective drift and diffusion of the sFML model. Left: recovery of the
    drift $a(x)$; Right: recovery of conditional standard deviation $b(x)$; (See \eqref{ab_lognormal} for the reference solution.)}
    \label{fig:ExpOU_dri_dif}
\end{figure}

Similar to the previous example, we can rewrite this SDE in the form of the classical SDE \eqref{Ito0} and obtain its true drift and diffusion,
\begin{equation} \label{ab_lognormal}
    \begin{split}
        a(x_n)&=\ln\left[\left(\mathbb{E}\left(\left.\frac{x_{n+1}}{x_n}\right|x_n\right)\right)^{1/\Delta}\right]=\ln(mx_n^{-\theta})+\frac{\sigma^2}{2},\\
        b(x_n)&=\text{Std}\left(x_{n+1}|x_n\right)=\sqrt{e^{\sigma^2\Delta}-1}(me^{\sigma^2/2})^\Delta (1-\theta\Delta)x_n.
    \end{split}
\end{equation}
On the other hand, from the learned sFML model, we can recover its effective drift and diffusion via
\begin{equation}
    \hat{a}(x) = \ln
    \left[\left(\mathbb{E}_z\left(\frac{\Gt_\Delta(x,z)}{x}\right)\right)^{1/\Delta}\right],
    \qquad \hat{b}(x) = \text{Std}_z(\Gt_\Delta(x,z)).
\end{equation}
The learned drift and diffusion are plotted in Figure \ref{fig:ExpOU_dri_dif}, where we observe excellent agreement with the reference solution.

\subsection{Multi-Dimensional Examples}

In this section, we present examples of learning multi-dimensional SDE systems. 

\subsubsection{Two-dimensional Ornstein–Uhlenbeck process}

We first consider a two-dimensional OU process
\begin{equation}\label{OU2d}
d \mathbf{x}_t=\mathbf{B} \mathbf{x}_t d t+\mathbf{\Sigma} ~ d \mathbf{W}_t,
\end{equation}
where $\mathbf{x}_t=\left(x_1, x_2\right) \in \mathbb{R}^2$ are the state variables, $\mathbf{B}$ and $\boldsymbol{\Sigma}$ are $(2 \times 2)$   matrices. In our test, we set
$$
\mathbf{B}=\left(\begin{array}{cc}
-1 & -0.5 \\
-1 & -1
\end{array}\right) \quad \boldsymbol{\Sigma}=\left(\begin{array}{cc}
1 & 0 \\
0 & 0.5
\end{array}\right)
$$
Our training data are generated with random initial conditions
uniformly sampled from $(-4,4)\times (-3,3)$, for a time up to $T=1.0$. For validation, we fix an initial condition of $\x_0=(0.3,0.4)$ and evolve the learned sFML model for up to $T=5$.

\begin{figure}[htbp]
  \centering
 
  \includegraphics[width=.43\textwidth]{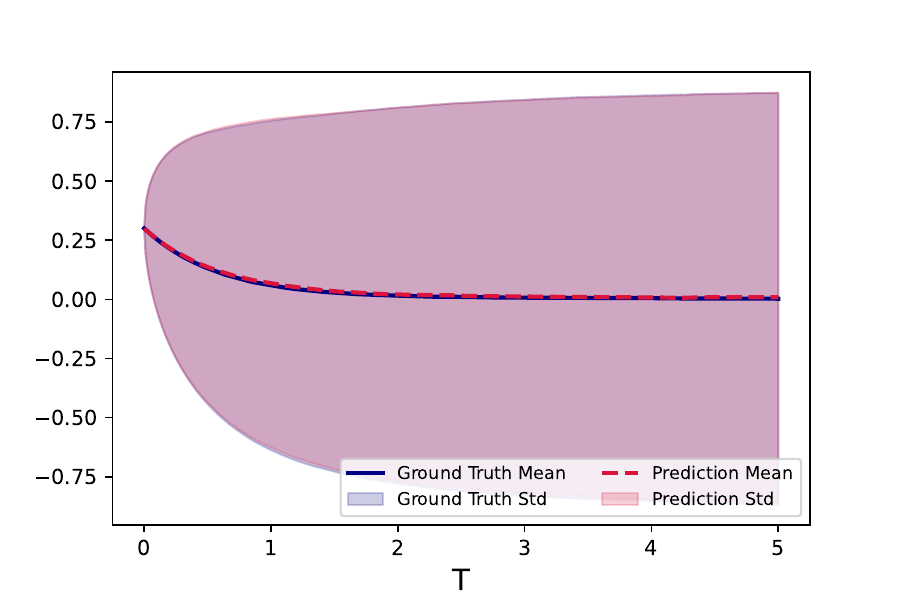}
  \includegraphics[width=.43\textwidth]{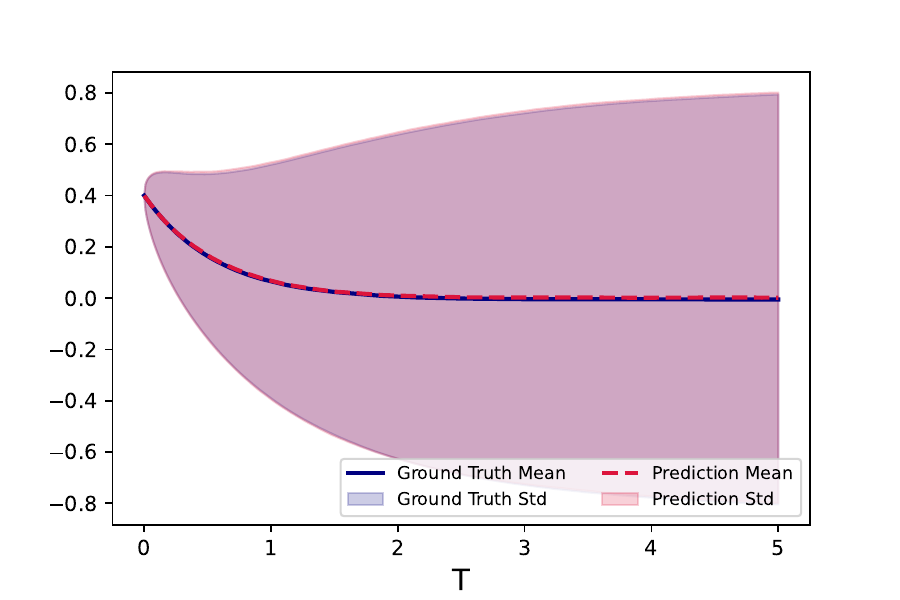}
  \caption{2D OU process \eqref{OU2d}: Mmean and standard deviation of the sFML model prediction with the initial condition $x_1=0.3, x_2 = 0.4$
    . Left: $x_1$; Right: $x_2$.}
     \label{fig:MdOU_show1}
\end{figure}

\begin{figure}[htbp]
  \centering
 
   \includegraphics[width=.43\textwidth]{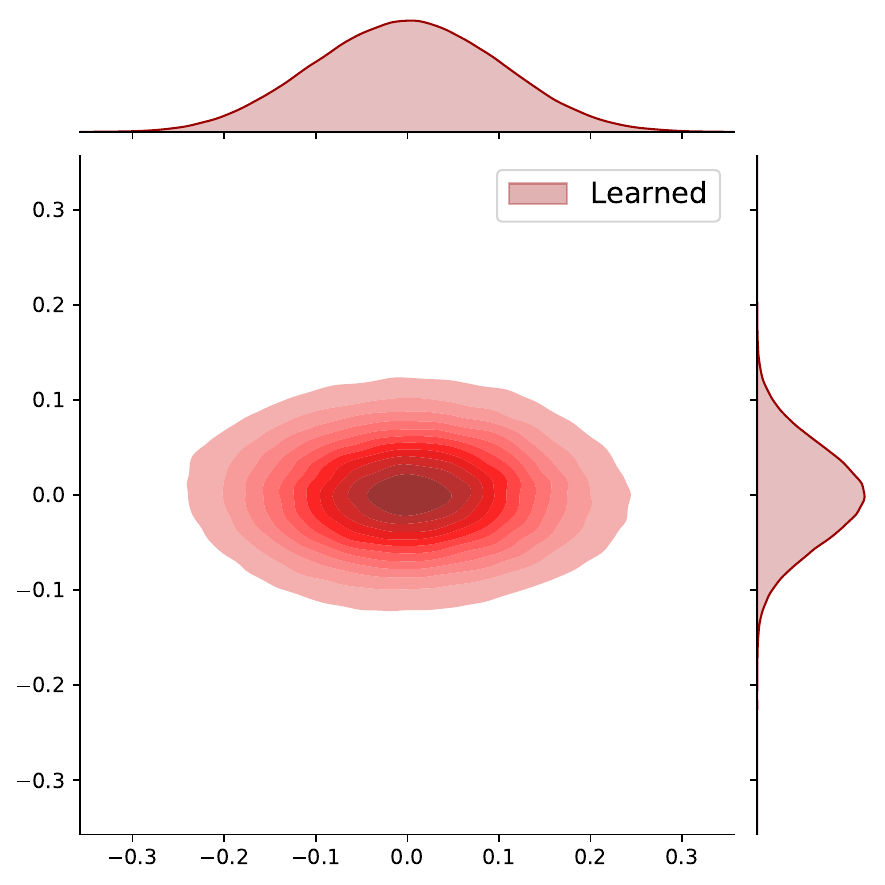}
 \includegraphics[width=.43\textwidth]{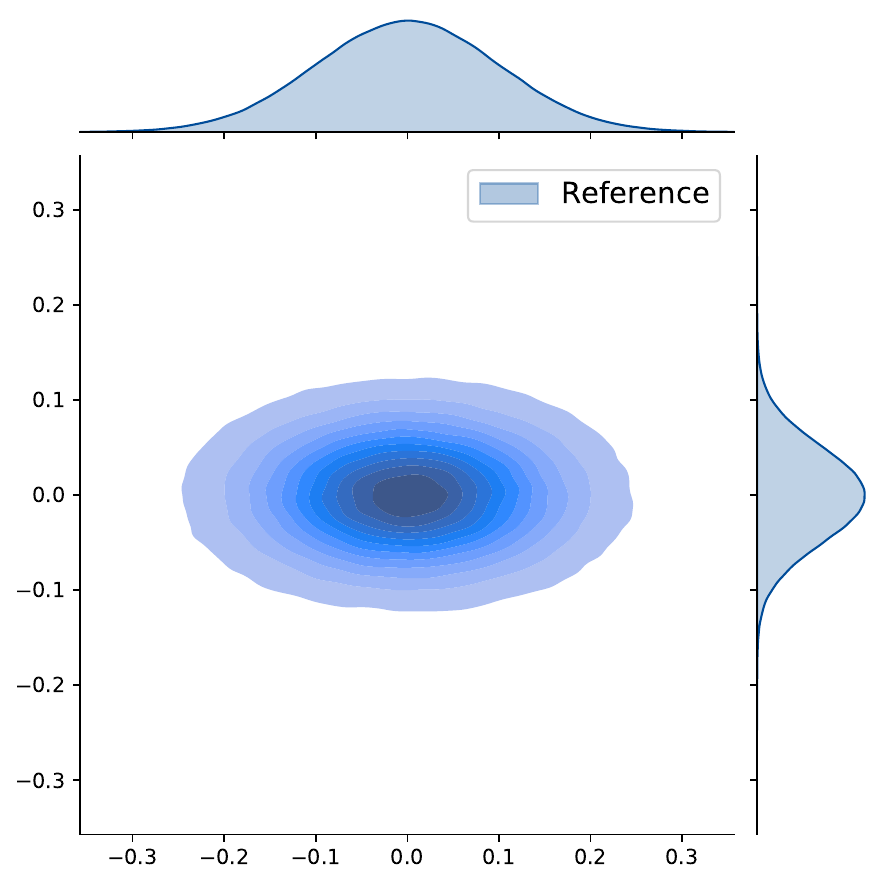}
  \caption{2D OU process \eqref{OU2d}: one-step conditional probability distribution from the learned decoder $\D_\Delta$ (left) and the ground truth (right), at $\x=(0,0)$.}
     \label{fig:MdOU_dis}
\end{figure}

The mean and the standard deviation of the sFML model predictions are shown in Figure \ref{fig:MdOU_show1}, where we observe very good agreement with the reference solutions.
To examine the conditional probability distribution generated by the learned sFML model, we present both the joint probability distribution and its marginal distributions, generated by the learned decoder $\D_\Delta(\x,\mathbf{z})$, $\z\sim N(\mathbf{0},\mathbf{I}_{2})$, at $\x=(0,0)$. The results are in Figure \ref{fig:MdOU_dis}, where good agreement with the true conditional distribution can be seen.

\subsubsection{Five-dimensional Ornstein–Uhlenbeck process}

We now consider a 5-dimensional OU process, driven by Wiener processes of varying dimensions
\begin{equation}\label{eq: OU5d}
d \mathbf{x}_t=\mathbf{B} \mathbf{x}_t d t+\mathbf{\Sigma} ~d \mathbf{W}_t,
\end{equation}
where $\mathbf{x}_t=\left(x_1, \ldots, x_5\right) \in \mathbb{R}^5$ are the state variables, $\mathbf{B}$ and $\boldsymbol{\Sigma}$ are $(5 \times 5)$ are matrices. While the $\mathbf{B}$ matrix is set as
$$
\mathbf{B}=\left(\begin{matrix}
     0.2 &     1.0 &     0.2 &     0.4 &     0.2\\
   -1.0 &     0.0 &     0.2 &     0.8 &   -1.0\\
     0.2 &     0.2 &   -0.8 &   -1.2 &     0.2\\
   -0.6 &     0.0 &     1.2 &   -0.2 &     0.6\\
     0.2 &     0.2 &     0.6 &     0.4 &     0.0
\end{matrix}\right),
$$
we choose the following 5 different cases for $\boldsymbol{\Sigma}$, whose ranks vary from  1 to 5. The objective is to examine the learning capability of the proposed sFML method when the underlying true stochastic dimension is unknown. The 5 cases we choose are:
$$
\mathbf{\Sigma_1}=\operatorname{diag} \left( 0,\, 0,\, 1,\, 0,\, 0\right)
$$
$$
\mathbf{\Sigma_2}=\operatorname{diag} \left( 0,\, 0.8,\, 0,\, 0,\, -0.8\right),\quad
\mathbf{\Sigma_3}=\left(\begin{matrix}
  0.8 &  0.2 &  0 &  0 &  0\\
 -0.4 &  0.6 &  0 &  0 &  0\\
  0 &  0 &  0 &  0 &  0\\
  0 &  0 &  0 &  0.7 &  0\\
  0 &  0 &  0 &  0 &  0
\end{matrix}\right),\,
$$

$$
\mathbf{\Sigma_4}=\left(\begin{matrix}
  0.7 &  0 & -0.4 &  0 &  0\\
  0 &  0 &  0 &  0 &  0\\
  0.1 &  0 &  0.6 &  0.2 & -0.1\\
  0 &  0 &  0.1 & -0.6 &  0.2\\
  0 &  0 &  0 &  0.3 &  0.8
\end{matrix}\right),
\mathbf{\Sigma_5}=\left(\begin{matrix}
  0.8 &  0.2 &  0.1 & -0.3 &  0.1\\
 -0.3 &  0.6 &  0.1 &  0 & -0.1\\
  0.2 & -0.1 &  0.9 &  0.1 &  0.2\\
  0.1 &  0.1 & -0.2 &  0.7 &  0\\
 -0.1 &  0.1 &  0.1 & -0.1 &  0.5
\end{matrix}\right),
$$
where the indices are chosen such that $\operatorname{rank}(\mathbf{\Sigma}_k) = k$, $k=1,\dots, 5$.

To generate the training data, we use random initial conditions drawn uniformly from the hypercube $(-4,4)^5$ and solve the system up to  $T=1.0$. For the test data, we set the initial condition to $\mathbf{x}_0 = \left(0.3,-0.2,-1.7,2.5,1.4\right)$ and evolve the learned sFML for up to $T=5$.

To construct the sFML model, a key parameter is the dimension $n_z$ of the latent variable $\z$. It needs to match the true dimension of the random process driven by the unknown stochastic system. Since the true random dimension is unknown, we propose to construct a sequence of sFML models, starting with a small random dimension $n_z$, for example, $n_z=1$. We then train more sFML models with progressively increasing values of $n_z$. During this process, we monitor the MSE losses \eqref{Loss2} of the models. The results of this procedure are tabulated in Table \ref{tab:training loss}. We observe that when the latent variable dimension $n_z$ matches the true random dimension, there is a significant drop of two to three orders in magnitude in the MSE losses. In practice, this can be used as a guideline to determine the proper dimension of the latent variable $\z$.
\begin{table}[tbhp]
\begin{center}
\begin{tabular}{|c|ccccc|} \hline
$n_z$ & $\mathbf{\Sigma}_1$ & $\mathbf{\Sigma}_2$ & $\mathbf{\Sigma}_3$ & $\mathbf{\Sigma}_4$ & $\mathbf{\Sigma}_5$ \\ \hline
  1& $3.5\times 10^{-7}$ &  $1.1\times 10^{-3}$ &  $1.8\times 10^{-3}$ &  $2.3\times 10^{-3}$ &  $3.3\times 10^{-3}$\\
  2& $5.1\times 10^{-7}$ &  $6.4\times 10^{-7}$ &  $7.2\times 10^{-4}$ &  $1.4\times 10^{-3}$ &  $1.9\times 10^{-3}$\\
  3& $3.3\times 10^{-7}$ &  $5.0\times 10^{-7}$ &  $7.1\times 10^{-7}$ &  $5.3\times 10^{-4}$ &  $8.8\times 10^{-4}$\\
  4& $3.8\times 10^{-7}$ &  $5.4\times 10^{-7}$ &  $4.1\times 10^{-7}$ &  $7.2\times 10^{-7}$ &  $4.5\times 10^{-4}$\\
  5& $7.2\times 10^{-7}$ &  $9.4\times 10^{-7}$ &  $6.3\times 10^{-7}$ &  $6.1\times 10^{-7}$ &  $6.2\times 10^{-7}$\\ \hline
\end{tabular}
\caption{5D OU process \eqref{eq: OU5d}: The MSE loss \eqref{Loss2} of the sFML models with latent variable dimension $n_z=1,\dots, 5$ with varying stochastic dimensions $k=1,\dots,5$, $\textrm{rank}(\mathbf{\Sigma}_k)=k$. Note the significant drops of the MSE losses from $n_z<k$ to $n_z\geq k$.}
\label{tab:training loss}
\end{center}
\end{table}

\begin{figure}[htbp]
  \centering
  \includegraphics[width=.9\textwidth]{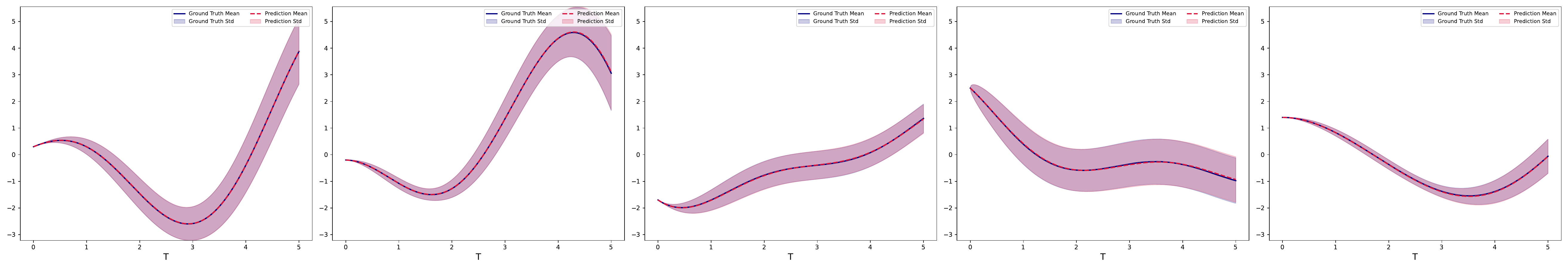}
  \includegraphics[width=.9\textwidth]{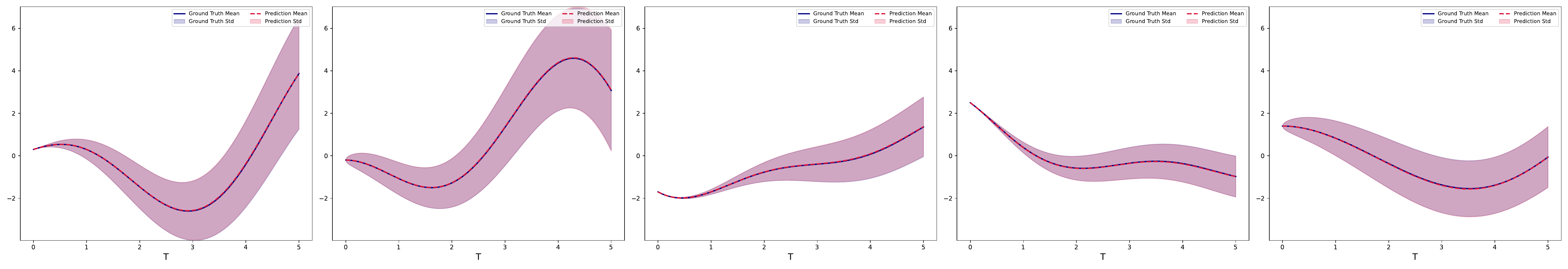}
  \includegraphics[width=.9\textwidth]{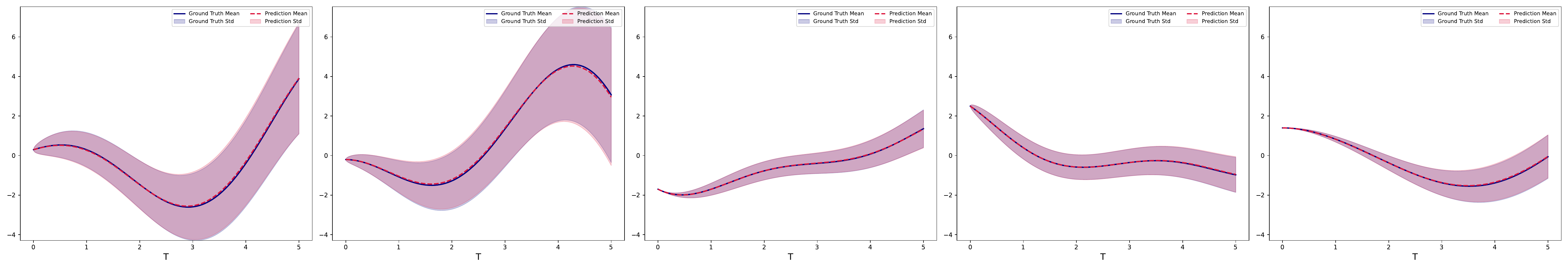}
  \includegraphics[width=.9\textwidth]{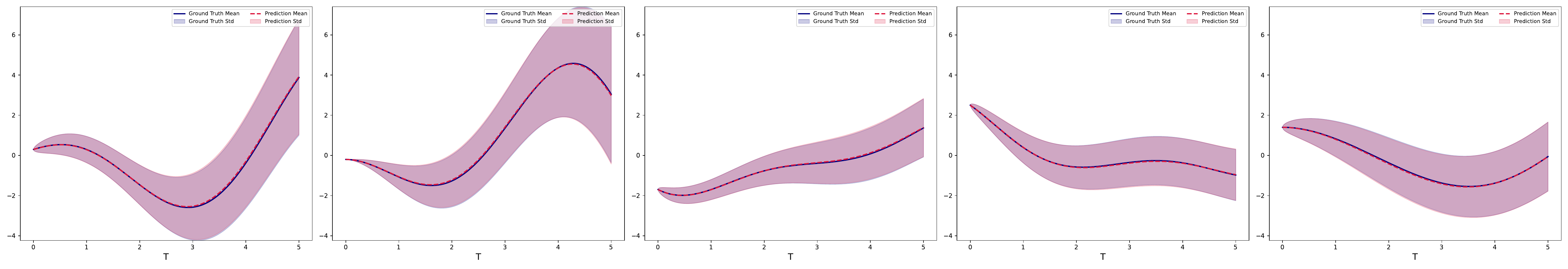}
  \includegraphics[width=.9\textwidth]{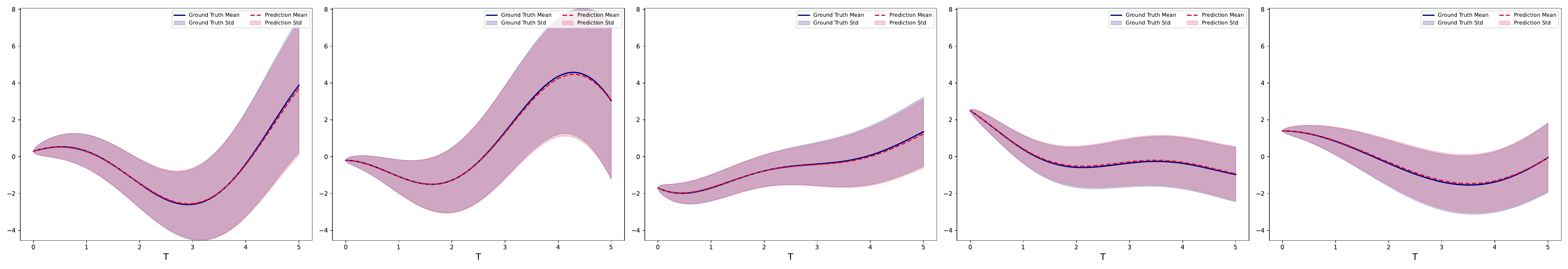}
    
  \caption{5D OU process \eqref{eq: OU5d}: Mean and standard deviation of the sFML model prediction for $x_1, \dots, x_5$ (left column to right), for the five different cases $\mathbf{\Sigma}_1, \dots, \mathbf{\Sigma}_5$ (top row to bottom).}
    \label{fig:Ex11OU5D meanstd}
\end{figure}

The learned sFML model predictions of the solution mean and standard deviation of each component are shown in Figure \ref{fig:Ex11OU5D meanstd}, for all five cases by using the matched latent variable dimension $n_z$. We observe excellent agreements with the reference solutions. 

\begin{figure}[htbp]
  \centering
  \includegraphics[width=.9\textwidth]{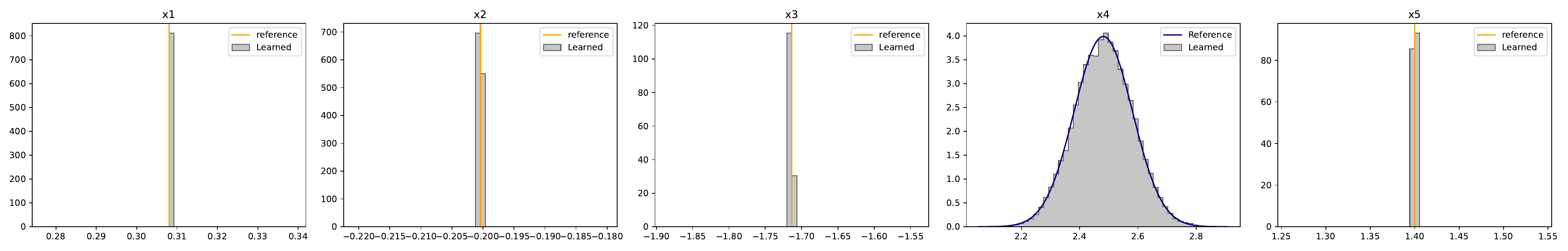}
  \includegraphics[width=.9\textwidth]{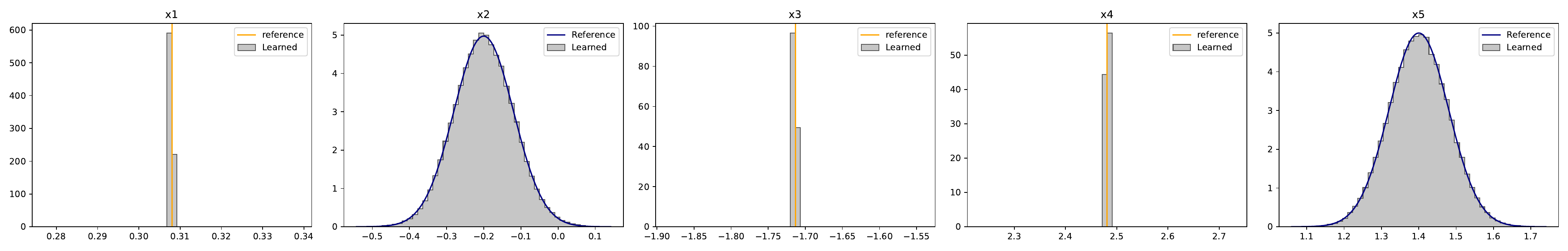}
  \includegraphics[width=.9\textwidth]{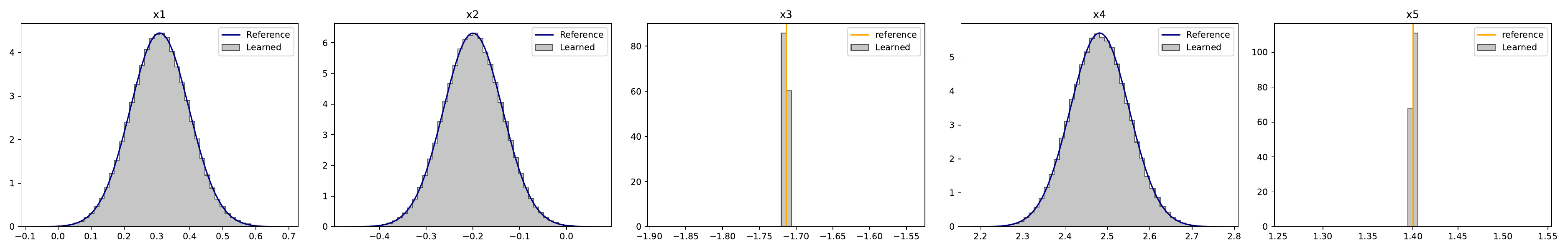}
  \includegraphics[width=.9\textwidth]{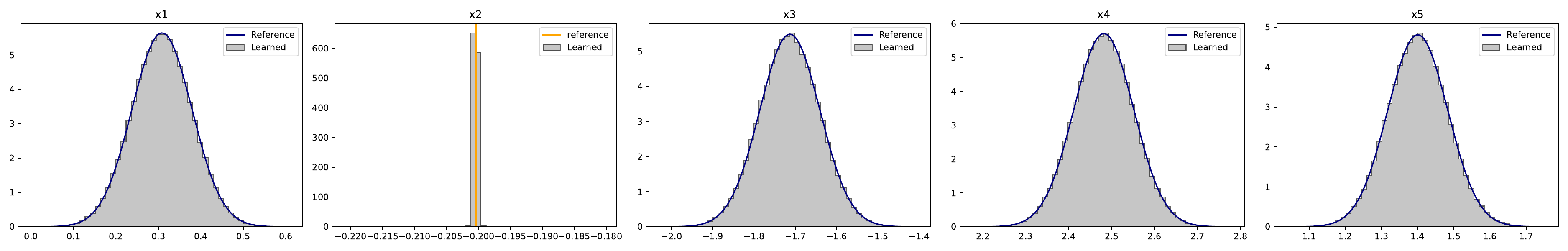}
  \includegraphics[width=.9\textwidth]{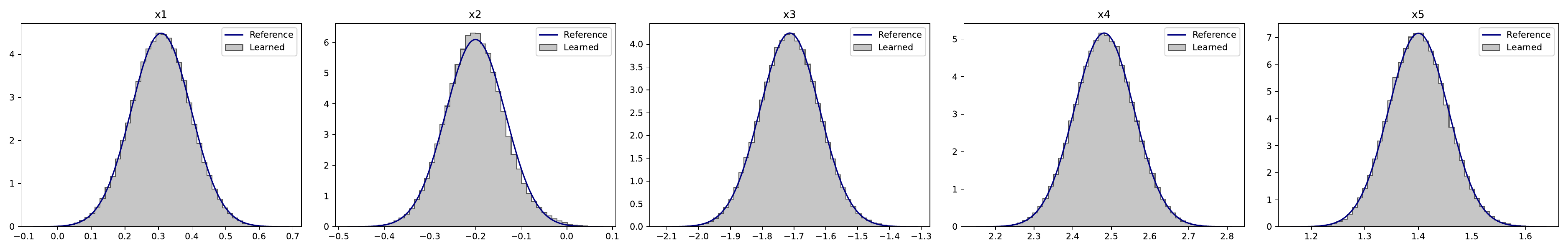}
    \caption{5D OU process \eqref{eq: OU5d}: One-step conditional probability distribution comparison: marginal distribution for $x_1, \dots, x_5$ (left column to right), for the five different cases $\mathbf{\Sigma}_1, \dots, \mathbf{\Sigma}_5$ (top row to bottom). Note the deterministic components with the Delta function distributions in each case.}
    \label{fig:Ex11OU5D 1step}
\end{figure}

For the one-step conditional probability distribution, we plot its marginal distributions for each of the components, $x_1, \dots, x_5$, obtained by the decoder of the learned sFML model, in Figure \ref{fig:Ex11OU5D 1step} (from left column to right), for each of the five cases $\mathbf{\Sigma}_1, \dots, \mathbf{\Sigma}_5$ (from top row to bottom).
We observe excellent agreement with the reference solutions. For the $k$-th case, $k=1,\dots, 5$, 
since $\operatorname{rank}(\mathbf{\Sigma}_k) = k$, the true random dimension is $k$. Consequently, there are $(5-k)$ non-random, i.e., deterministic, components. We clearly observe this in the marginal distributions, where the deterministic components exhibit Delta function distribution. We observe that the learned sFML models accurately reflect this and produce the correct Delta function distributions in each of the five cases.
\section{Conclusion}
\label{sec:conclu}
In this paper, we presented a numerical method for modeling unknown stochastic dynamical systems using trajectory data. The method is based on learning the underlying stochastic flow map, which is parameterized by Gaussian latent variables via an autoencoder architecture. The encoding function recovers the unobserved Gaussian random variables and the decoding function reconstructs the trajectory data. Loss functions are carefully designed to ensure the autoencoder sFML model accurately learns the stochastic components of the unknown system. By using a comprehensive set of numerical examples, we demonstrate that the proposed sFML model is highly effective and able to produce long-term system predictions well beyond the time horizon covered by the training data sets. The method can also automatically identify the true stochastic dimension of the unknown system. This aspect is important for practical applications and will be studied more rigorously in future work, beyond the academic example investigated in this paper.

\bibliographystyle{siamplain}
\bibliography{references}
\end{document}